\documentclass[runningheads]{llncs}

\usepackage{eccv}

\usepackage{eccvabbrv}

\usepackage{graphicx}
\usepackage{booktabs}

\usepackage{makecell}
\usepackage{array}
\usepackage{multirow}
\usepackage[table,xcdraw]{xcolor}
\usepackage[ruled,vlined,linesnumbered]{algorithm2e} %
\usepackage{amsmath,amssymb}

\usepackage{colortbl} %

\usepackage{caption}
\usepackage{subcaption}
\usepackage{hyperref}

\usepackage{orcidlink}

\hypersetup{hidelinks}
\begin{document}

\title{Unified Prediction and Planning via Conflict-Aware Disjoint Parameter Training}

\titlerunning{Conflict-Aware Disjoint Parameter Training}

\author{Taewon Seo\inst{1}$^\star$\orcidlink{0009-0007-1022-3354} \and
Seonae Jeon\inst{1}$^\star$\orcidlink{0009-0006-4452-3542} \and
Giwon Lee\inst{2}$^\star$\orcidlink{0009-0006-5472-2078} \and
Kuk-Jin Yoon \inst{2}$^\dagger$\orcidlink{0000-0002-1634-2756} \and
Daehee Park \inst{1}$^\dagger$\orcidlink{0000-0002-3961-6932}}

\authorrunning{Seo et al.}

\institute{
DGIST, Republic of Korea\\
\email{\{taewonseo,seonae,dhpark\}@dgist.ac.kr}
\and
KAIST, Republic of Korea\\
\email{\{dlrldnjs,kjyoon\}@kaist.ac.kr}
}

\maketitle

\begingroup
\renewcommand\thefootnote{}
\footnotetext{
$^\star$ Equal contribution. \quad
$^\dagger$ Corresponding authors.
}
\endgroup

\begin{abstract}
Accurate motion prediction of surrounding agents and safe motion planning are two closely coupled key tasks for social robot navigation in crowded environments.
Deploying these systems on resource-constrained edge devices necessitates compact, unified models that can perform both tasks simultaneously.
However, within these compact shared encoders, recent unified models often overlook severe representational conflicts that arise from the distinct objectives of predicting neighbor behaviors versus ego-centric safety planning.
To address this issue, we first identify the \textit{Skill Conflict}—a phenomenon where overlapping parameter assignments cause distinct tasks to compete for the same weights, preventing the model from fully specializing in individual skills.
To resolve this, we propose a novel model-merging-based framework, \textbf{Disjoint Parameter Training (DPT)}.
DPT mitigates performance degradation caused by Skill Conflict through distributed parameter learning, which separates the key parameter regions of each task while preserving their core capabilities prior to merging.
In addition, we observe that sparse merging, which selectively integrates only the most influential parameters for each task rather than combining all task-specific parameters, yields optimal performance by preventing interference among adjacent features and concentrating representational capacity.
DPT can be applied in parallel with a variety of merging methods.
Evaluated on standard crowd navigation benchmarks (JRDB and JTA), our framework demonstrates superior performance, validating its versatility and effectiveness for safe, resource-efficient robot navigation. The project page is available at: \url{https://dpt2026.github.io/}
\keywords{Motion Prediction \and Motion Planning \and Model Merging.}
\end{abstract}

\section{Introduction}
\label{sec:intro}

Motion planning~\cite{sadat2020perceive_intro_planning1, zeng2020dsdnet_intro_planning2,chen2024end_intro_planning3,yurtsever2020survey_intro_planning4} aims to generate future trajectories of an ego agent based on the surrounding environment and the behaviors of nearby agents.
For safe and socially compliant navigation in highly interactive scenarios, such as mobile robots navigating through dense crowds, the agent must accurately predict how humans will move and plan its own trajectory accordingly~\cite{E2E_4_ICCV2025_tang2025hip}.
Therefore, achieving both accurate prediction and safe planning simultaneously within a constrained computational budget is essential for reliable and user-friendly mobile service robots.

\begin{figure}[t]
    \centering
    \includegraphics[width=0.98\linewidth]{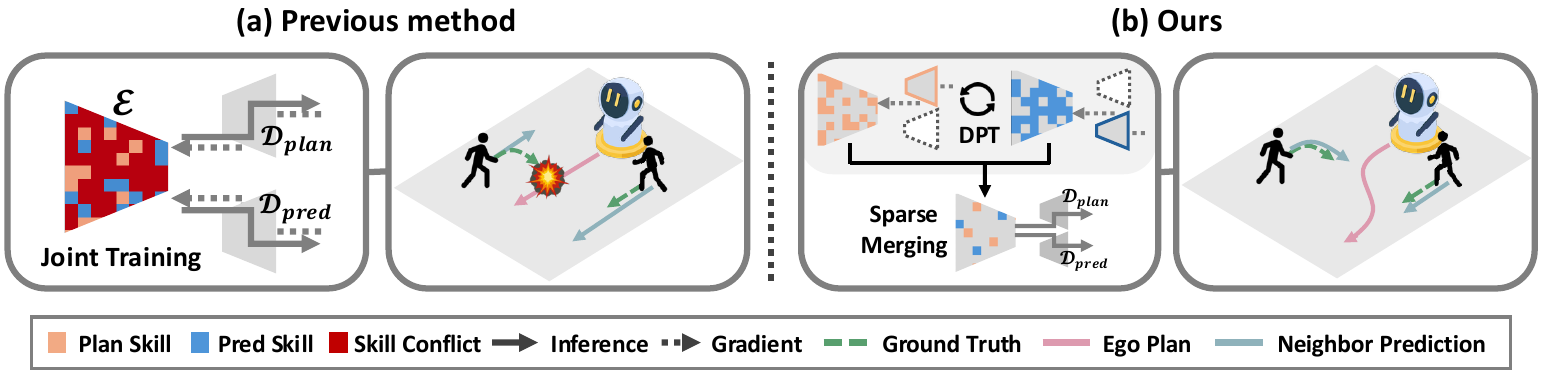}
    \caption{Comparison of learning strategies in unified models. (a) In previous methods, the skills of each task are learned within a single shared encoder, causing conflicts between the two skills in the parameter space.
    (b) DPT resolves this issue by separately training the major parameter regions allocated to each task and Sparse Merging combines only their core parameter regions, thereby mitigating skill conflicts within the shared encoder.
    }
    \label{fig:intro}
\end{figure}

To realize this efficiently, recent research has focused on unified frameworks~\cite{kedia2023GameTheoretic,huang2023DIPP,huang2024DTPP,huang2023gameformer} that jointly perform prediction and planning, replacing traditional sequential pipelines~\cite{note2eplan_1_ICCV2025_zhong2025robotrom, TP_49_old_zhou2023query, TP_11_ICCV2025_pei2025foresight,TP_13_ICCV2025_li2025natra}.
Such unified models~\cite{huang2023gameformer,huang2023DIPP,huang2024DTPP} typically use a shared encoder that serves as the representational backbone for both tasks, as shown in Fig.~\ref{fig:intro} (a).
While these architectures achieve strong joint reasoning performance, they generally overlook representational conflicts that arise in the shared encoder, especially when the model capacity is limited for edge deployment.
Prediction requires representations sensitive to the varied behaviors and intentions of surrounding agents~\cite{TP_intention_1_shi2022motion, TP_intention_2_CVPR2025_chen2025socialmoif, TP_intention_3_ICLR2025_fangneuralized, TP_intention_4_old_zhou2022hivt}, whereas planning demands ego-centric features that prioritize safety and feasibility~\cite{E2E_27_ICRA2025_yang2025uncad,E2E_19_NeurIPS2025_shang2025drivedpo,E2E_20_NeurIPS2025_zhou2025autovla, E2E_18_NeurIPS2025_chen2025temporal}.
Optimizing both objectives on the same compact encoder leads to conflicting gradient updates, which can severely degrade the feature quality of both tasks~\cite{merging_TaskConf_GradConf_1_ICLR2025_sun2025task, merging_LLM_TaskConf_GradConf_2_ICLR2025_yoshida2025mastering, merging_LLM_TaskConf_GradConf_3_CVPR2024_workshop_li2024improving, merging_LLM_TaskConf_GradConf_1_ICLR2025_lee2025mitigating}.

To address this challenge without simply inflating model size, we propose a model-merging-based unified framework that explicitly mitigates task interference in the shared encoder, as shown in Fig.~\ref{fig:intro} (b).
Model merging~\cite{Merging_Survey_yang2024model} is a methodology that combines the parameters of models fine-tuned for individual tasks to create a single model that can effectively perform all tasks simultaneously.
Unlike ensemble learning~\cite{li2022ensemble, arpit2022ensemble} or mixture-of-experts~\cite{MoE_xu2025limoe,MoE_he2025capacity,MoE_jiang2025expertad,MoE_huang2025discovering}, which handle multiple objectives by maintaining multiple models or expert branches, model merging produces a single model, making it effective for autonomous driving where inference latency is critical and both objectives must be achieved within a single deployment.
Existing model merging methods, however, have primarily been studied in the context of large-scale models (e.g., LLMs) with massive parameter counts and have mainly focused on language-related tasks~\cite{merging_LLM_1_NIPS2025_li2025model,merging_LLM_2_NIPS2025_bercovich2025ffn,merging_LLM_3_NIPS2025_theus2025generalized,merging_LLM_4_ICLR2025_jiang2025fine,merging_LLM_TaskConf_1_NIPS2025_yang2025continual,merging_LLM_TaskConf_2_NIPS2025_panariello2025accurate,merging_LLM_TaskConf_3_NIPS2025_zeng2025robustmerge}.
Therefore, a significant gap exists between these approaches and resource-constrained robot navigation settings, which rely on compact unified models.
This discrepancy induces a critical issue during the merging process: in models with limited capacity, the parameter regions crucial for each task tend to densely overlap and conflict with one another.
When the overall model size is small, each task must utilize a larger portion of the shared parameters to achieve strong single-task performance, which inevitably intensifies this interference.

We define this phenomenon as the \textit{Skill Conflict}, a condition where overlapping parameter assignments cause distinct tasks to compete for the same weights, preventing the model from fully specializing in individual skills.
To resolve this, we propose \textbf{Disjoint Parameter Training (DPT)}, which fine-tunes task-specific models by distributing the parameter learning regions across tasks, and then merges them into a single compact unified model.
DPT separates the learning regions of each task by exchanging model-sized binary masks, each indicating a unique parameter region assigned to that task.
Moreover, instead of allocating all parameter regions for training from the beginning, it gradually expands the learnable parameter space from a local region, allowing each task’s core skill to be compactly represented within a small parameter area.
Building on this property, we further apply sparse merging, which selectively integrates only the most influential parameters from each task rather than merging all, enabling more efficient and skill-focused unification of the shared encoder while preventing adjacent feature interference.
This design enables each task to specialize in distinct subspaces of the shared representation, preserving both prediction accuracy and planning safety within a limited model capacity.

Our main contributions are summarized as follows:
\begin{itemize}
    \item We define and analyze the \textit{Skill Conflict} problem caused by shared encoder representations in compact, unified prediction-planning models for robot navigation.
    \item We propose Disjoint Parameter Training (DPT), which separates and localizes each task’s core skills within distinct parameter regions, and we introduce sparse merging to prevent representational interference among neighboring features.
    \item Our method achieves superior performance, maintaining accurate prediction and safe planning across standard crowd navigation benchmarks (JRDB and JTA), validating its effectiveness for resource-efficient mobile robots.
\end{itemize}

\section{Related Works}
\label{sec:related_works}

\subsection{Unified Motion Prediction and Planning}
Motion prediction and motion planning are the key tasks for achieving safe navigation in mobile robots and autonomous systems.
Motion prediction forecasts the future trajectories of surrounding agents based on current and past information~\cite{TP_1_CVPR2025_taketsugu2025physical, TP_2_CVPR2025_dong2025leveraging, TP_3_CVPR2025_yang2025tra, TP_4_CVPR2025_messaoud2025towards, TP_5_CVPR2025_yu2025enduring, TP_6_CVPR2025_bahari2025certified, TP_8_CVPR2025_qiu2025adapting, TP_9_ICCV2025_ren2025totp}, while motion planning generates the ego agent’s trajectory using these predictions and scene context~\cite{note2eplan_2_ICCV2025_xu2025daa, note2eplan_3_ICCV2025_lee2025interaction, note2eplan_4_ICLR2025_zheng2025diffusion, note2eplan_5_CVPR2024_yang2024diffusion, note2eplan_6_ICCV2023_wang2023dreamwalker}.
Conventional sequential pipelines predict first and plan later~\cite{note2eplan_7_IROS2025_lee2025non,note2eplan_8_CVPR2025_zhang2025carplanner,TP_10_ICCV2025_park2025generative, TP_11_ICCV2025_pei2025foresight, TP_12_ICCV2025_knoche2025donut}, training modules independently~\cite{TP_13_ICCV2025_li2025natra, TP_14_ICCV2025_rao2025amd, TP_15_NeurIPS2025_wendriving, TP_16_ICLR2025_suninteractive, TP_19_ICLR2025_akbiyikleveraging}, which increases real-time inference latency and prevents learning interdependent task relationships~\cite{E2E_4_ICCV2025_tang2025hip, E2E_25_ICLR2025_jia2025drivetransformer,E2E_41_old_weng2024drive_endtoend5,li2025recogdrive,jiang2026vadv,su2026drivemamba,li2025discrete}.
To address these bottlenecks, recent studies have developed unified models that jointly reason over planning and prediction within a single framework~\cite{huang2023DIPP, huang2024DTPP, huang2023gameformer}.
However, these prior studies often rely on scaling up model parameters, overlooking the representational task conflict that arises in the shared encoder due to the distinct objectives of prediction and planning.
This conflict becomes particularly fatal in compact models deployed on resource-constrained edge devices.
In this work, we verify the existence of such task conflicts in unified models and propose a framework that mitigates them through task-wise distributed learning, ensuring high performance for both tasks.

\subsection{Model Merging}
Model merging~\cite{Merging_Survey_yang2024model} combines the parameters of distinct fine-tuned models into a single architecture to perform multiple tasks simultaneously.
While early approaches relied on simple or weighted parameter averaging~\cite{Merging_Average1_izmailov2018averaging,Merging_Average2_wortsman2022model,Merging_Fisher_matena2022merging}, direct combination often increases task interference and degrades performance.
To mitigate this, Task Arithmetic~\cite{Merging_Taskarithmetic_ilharco2022editing} uses task vectors to preserve directional characteristics, Ties Merging~\cite{Merging_Tiesmerging_yadav2023ties} resolves sign conflicts via majority voting, and Localize-and-Stitch~\cite{Merging_localizeandstich_he2024localize} learns local masks to isolate critical parameter regions.
Despite these advancements in reducing task conflicts~\cite{Merging_DAREmerging_yu2024language,Merging_Dellamerging_deep2024della,merging_TaskConf_1_ICLR2025_liu2025skill, merging_TaskConf_2_CVPR2024_xu2024training, merging_TaskConf_3_CVPR2024_yu2024boosting}, most existing methods are tailored exclusively for Large Language Models (LLMs) equipped with massive parameter capacities~\cite{merging_LLM_TaskConf_4_NIPS2025_sun2025towards,merging_LLM_TaskConf_5_NIPS2025_yang2025mix,merging_LLM_TaskConf_6_NIPS2025_zhou2024hm3,merging_LLM_TaskConf_7_ICLR2025_wang2024lines}.
These highly parameterized settings differ fundamentally from our target domain, which requires deploying compact unified models on resource-constrained edge devices for robot prediction and planning.
In such compact architectures, parameter overlap inevitably leads to severe representational interference.
Therefore, we identify and analyze this \textit{Skill Conflict} and propose the Disjoint Parameter Training (DPT) framework to actively prevent parameter collision, successfully maximizing multi-task performance within a strictly limited parameter budget.

\section{Methods}
\label{sec:methods}

We aim to train a unified model that jointly handles motion prediction and motion planning.
Among the following sections, Sec.~\ref{sec:problem_definition} provides the problem definition, describing the objective we aim to achieve.
Sec.~\ref{sec:prelim_merging} introduces preliminaries on model merging and task vectors, and clarifies why naive fine-tuning and merging, which work reasonably well for large foundation models, are problematic for compact unified models.
Sec.~\ref{sec:skill_conflict} provides empirical evidence of \textit{Skill Conflict} in a shared-encoder prediction–planning setting.
Sec.~\ref{sec:DPT} introduces \textit{Disjoint Parameter Training (DPT)}.
This fine-tuning strategy prepares task-specialized \textit{material} models, which act as individual fine-tuned ingredients for the merging process, with strictly non-overlapping core regions.
Finally, Sec.~\ref{sec:sparse_merging} describes our \textit{Sparse Merging} strategy that combines these material models by using only salient task-vector components.

\subsection{Problem Definition}
\label{sec:problem_definition}
We consider a unified architecture with a shared encoder and two task heads: a prediction decoder and a planning decoder.
Let the input space be $\mathbb{X}$ and the output space be $\mathbb{Y}$, where $X \in \mathbb{X}$ and $Y \in \mathbb{Y}$.
In a scene with $N$ agents, agent $0$ denotes the ego, and agents $1:N-1$ are surrounding agents.

To capture rich human behaviors, the multi-modal input incorporates human trajectories, poses, and bounding boxes, denoted as $X_{\text{traj}}$, $X_{\text{pose}}$, and $X_{\text{box}}$, respectively.
The unified model input is defined as:
\begin{equation}
    X = \left\{
    X_{\text{traj},\,0:N-1}^{-T_{\mathrm{obs}}:0},\
    X_{\text{pose},\,0:N-1}^{-T_{\mathrm{obs}}:0},\
    X_{\text{box},\,0:N-1}^{-T_{\mathrm{obs}}:0}
    \right\}.
\end{equation}
The output is defined as:
\begin{equation}
    Y = \left\{ Y_{\mathrm{plan}},\ Y_{\mathrm{pred}} \right\},
\end{equation}
where the planned future ego trajectory is
\begin{equation}
    Y_{\mathrm{plan}} = \left\{ Y_{0}^{1:T_{\mathrm{fut}}} \right\},
\end{equation}
and the predicted future trajectories of neighbors are
\begin{equation}
    Y_{\mathrm{pred}} = \left\{ Y_{1:N-1}^{1:T_{\mathrm{fut}}} \right\}.
\end{equation}
Here, $T_{\mathrm{obs}}$ is the number of observed steps and $T_{\mathrm{fut}}$ is the prediction/planning horizon.
Our goal is to learn the shared parameters so that the unified model can produce safe $Y_{\mathrm{plan}}$ and accurate $Y_{\mathrm{pred}}$.
We specifically focus on mitigating representational interference within the shared encoder, a critical bottleneck when determining parameter assignments in compact models.

\subsection{Preliminaries: Model Merging and Task Vectors}
\label{sec:prelim_merging}
Model merging begins by preparing task-specialized “material” models, and then combines them into a unified model.
Let $\Theta_0$ be a pretrained backbone and $t \in {T}=\{\text{pred},\text{plan}\}$ denote the target tasks.
In the \textit{naive material preparation}, each task is fine-tuned \emph{independently} from the same initialization $\Theta_0$, with no parameter or gradient sharing across tasks:
\begin{equation}
\Theta_t=\arg\min_{\Theta}\mathbb{E}_{(X,Y)\sim \mathcal{D}_t}\big[\mathcal{L}_t(f_{\Theta}(X),Y)\big].
\end{equation}
The change induced by independent fine-tuning defines the \textit{task vector}~\cite{Merging_Taskarithmetic_ilharco2022editing}:
\begin{equation}
\tau_t := \Theta_t - \Theta_0 .
\end{equation}
After preparing ${\Theta_t}$ (or equivalently ${\tau_t}$), a merged model is produced by an operator that combines task contributions:
\begin{equation}
\Theta_{\mathrm{merge}}=\mathcal{M}\left(\Theta_0,{\tau_t},\phi\right),
\end{equation}
where $\mathcal{M}$ denotes a chosen merging family (e.g., scalar-weighted linear combination, coordinate-wise masking, or learned gating) and $\phi$ denotes the merging coefficient associated with the operator $\mathcal{M}$.

For massive foundation models, naive per-task fine-tuning followed by merging often works reasonably well, as their massive capacity naturally allocates distinct skills into separate, sparse parameter subspaces.
This was observed in the prior work~\cite{Merging_localizeandstich_he2024localize}, which found that utilizing only the top 10\% (or even 1\%) of a task vector can fully preserve single-task performance in large models.
However, in compact unified models designed for edge robots, parameter capacity is strictly limited.
Consequently, independent fine-tuning forces different tasks to update densely overlapping regions.
This overlap causes interference when models are merged, particularly within the shared encoder that both tasks rely on.
We find that the other merging methods in Tab.~\ref{tab:MainTable1} do not yield effective performance on the compact unified model, which supports our claim.

\subsection{Empirical Evidence of Skill Conflict in a Shared Encoder}
\label{sec:skill_conflict}
We begin by defining a binary mask of the same size as the model parameters to specify how much of a task vector is utilized.
For task $t \in T=\{{\mathrm{pred},\mathrm{plan}}\}$ and utilization ratio $K$, the \textit{activation mask} $M_t^{(K)}$ selects the top-$K$ coordinates of the task vector by magnitude:
\begin{equation}
M_t^{(K)} = \mathbb{I}\!\big(|\tau_t| \ge \lambda_{K\%}(|\tau_t|)\big),
\end{equation}
where $\lambda_{K\%}(\cdot)$ is the $K$-th percentile of $|\tau_t|$. We refer to $K$ as the \textit{mask ratio}, representing the task-vector utilization percentage.
By adjusting this \textit{mask ratio}, we can control the capacity allocated to each task and systematically observe how differing levels of parameter usage impact both individual and joint performance.

\begin{figure}[t]
    \centering
    \begin{minipage}[t]{0.49\linewidth}
        \centering
        \includegraphics[width=\linewidth]{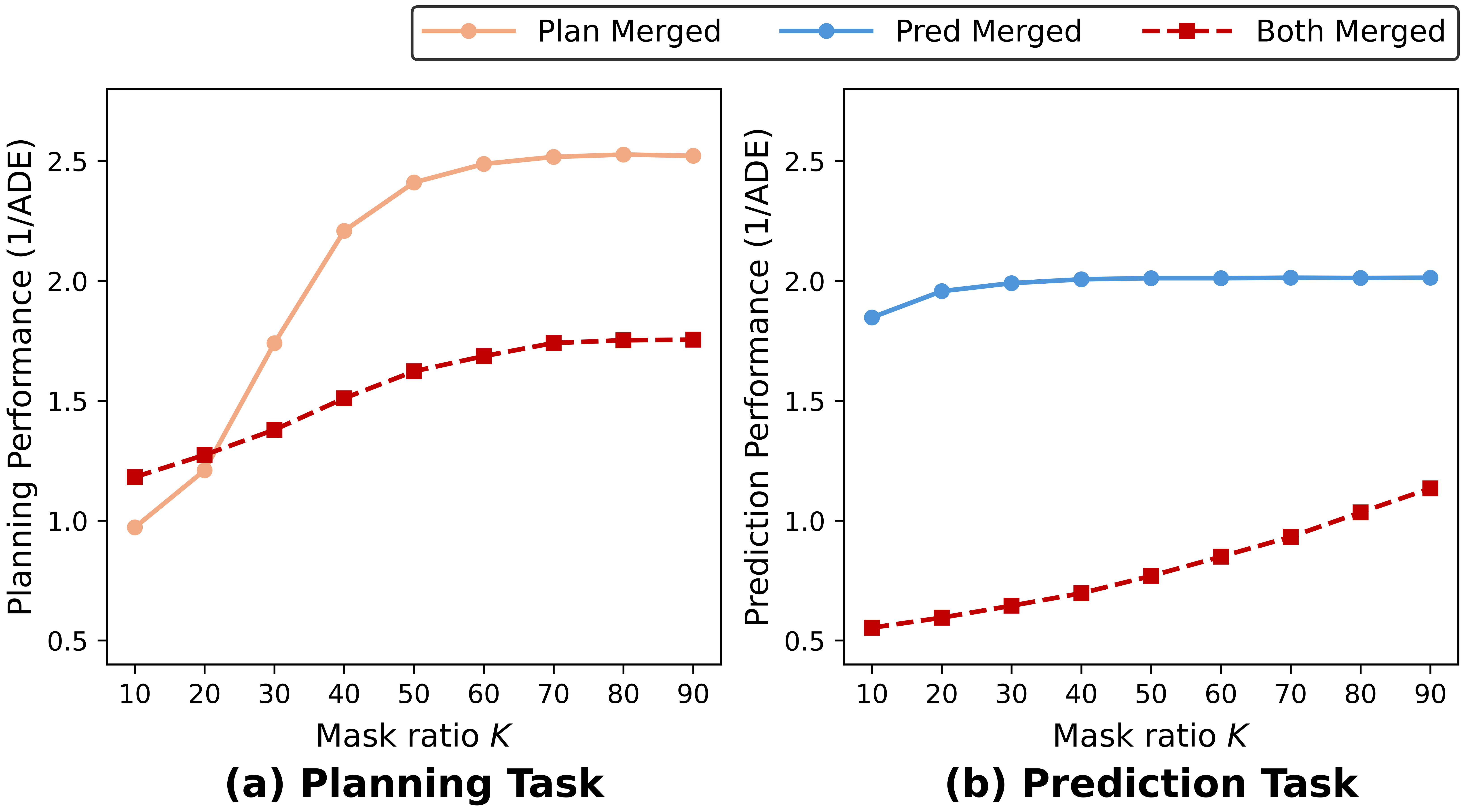}
        \caption{Planning and Prediction performance according to the \textit{mask ratio} $K$. \textit{Plan Merged} and \textit{Pred Merged} use the single task vectors corresponding to the mask ratio. In \textit{Both Merged}, the parameter utilization of the non-target task is fixed at 40\%, while that of the target task is used according to the given parameter ratio.}
        \label{fig:fig_3_ordinary_perf_graph}
    \end{minipage}\hfill
    \begin{minipage}[t]{0.49\linewidth}
        \centering
        \includegraphics[width=\linewidth]{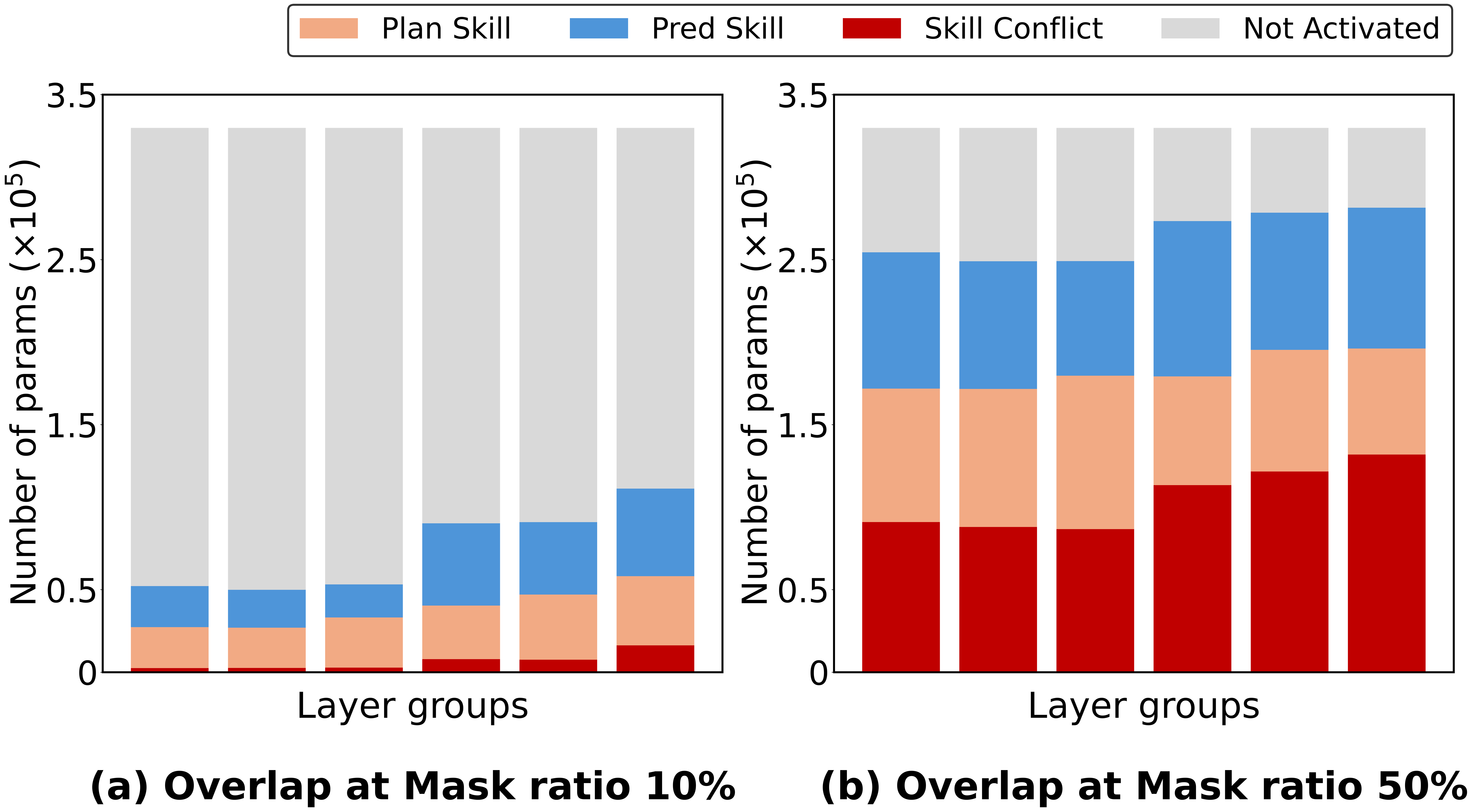}
        \caption{Parameter distribution by layer group and the degree of \textit{Skill Conflict} between \textit{Plan} and \textit{Pred Skill} at different mask ratios. Both tasks use an equal proportion of task vectors for the given mask ratio. At (a) 10\% mask ratio, little \textit{Skill Conflict} occurs, whereas at (b) 50\% ratio, significant conflict is observed.}
        \label{fig:fig_4_ordinary_overlap_bar}
    \end{minipage}
\end{figure}

To understand how capacity limitations affect multi-task merging, we first investigate the relationship between parameter utilization and task performance using Fig.~\ref{fig:fig_3_ordinary_perf_graph}.
This figure sweeps the \textit{mask ratio} $K\!\in\![10\%,90\%]$ and reports the performance ($1/\text{ADE}$; higher is better).
For each $K$, we evaluate single-task merges (\textit{Plan Merged}: $\Theta_0 + M_{\text{plan}}^{(K)} \odot \tau_{\text{plan}}$, \textit{Pred Merged}: $\Theta_0 + M_{\text{pred}}^{(K)} \odot \tau_{\text{pred}}$) and a two-task merge (\textit{Both Merged}: $\Theta^{(K,40\%)} \;=\; \Theta_0 \;+\; M_{\text{target}}^{(K)} \odot \tau_{\text{target}} \;+\; M_{\text{other}}^{(40\%)} \odot \tau_{\text{other}}$), where the non-target task is held at a constant 40\% utilization.
In the single-task cases, performance steadily improves as $K$ increases and plateaus at a high level, confirming that utilizing a larger portion of the task vector is beneficial when merged alone.
Conversely, in the two-task setting, attempting to increase $K$ yields only marginal gains before rapidly saturating at a substantially lower ceiling.
This early saturation demonstrates that greedy parameter utilization by one task severely degrades performance of the other task.

To diagnose the underlying cause of this early saturation, Fig.~\ref{fig:fig_4_ordinary_overlap_bar} visualizes how the prediction and planning tasks allocate parameters within the shared encoder.
We categorize the parameters into four groups based on their activation masks $M_{\text{plan}}^{(K)}$ and $M_{\text{pred}}^{(K)}$: \textit{Plan Skill}, \textit{Pred Skill}, \textit{Skill Conflict} (activated by both), and \textit{Not Activated}.
We then quantify the exact coordinate-wise collision by defining the overall overlap as:
\begin{equation}
\mathrm{Overlap}(K) \;=\; \frac{1}{D}\sum_{j=1}^{D}\big(M_{\text{pred}}^{(K)}[j]\land M_{\text{plan}}^{(K)}[j]\big)\times 100\% ,
\end{equation}
where $D$ is the total number of parameters activated by at least one task.
The analysis reveals that at a conservative mask ratio ($K{=}10\%$), the tasks activate small, largely distinct regions with minimal conflict (Overlap $8.35\%$).
However, as both tasks expand their utilization to $K{=}50\%$, their activated regions heavily overlap within the same layer groups, causing the overlap metric to surge to $39.25\%$.
Thus, the more we try to make the model stronger for each individual task by utilizing larger parameter regions, the more entangled the shared encoder becomes, which explains the early saturation of \textit{Both Merged} in Fig.~\ref{fig:fig_3_ordinary_perf_graph}.

All empirical studies are conducted on the Social-Transmotion~\cite{TP_39_ICLR2024_saadatnejad2023social} backbone, with detailed architectural specifications provided in Supplementary Section 2.
We also conduct a similar analysis on DTPP~\cite{huang2024DTPP} (Supplementary Section 6), showing that our findings are common across different backbones.
This motivates preparing material models that minimize overlap (Sec.~\ref{sec:DPT}) and merging them sparsely (Sec.~\ref{sec:sparse_merging}).

\begin{figure*}[t]
    \centering
    \includegraphics[width=0.99\textwidth]{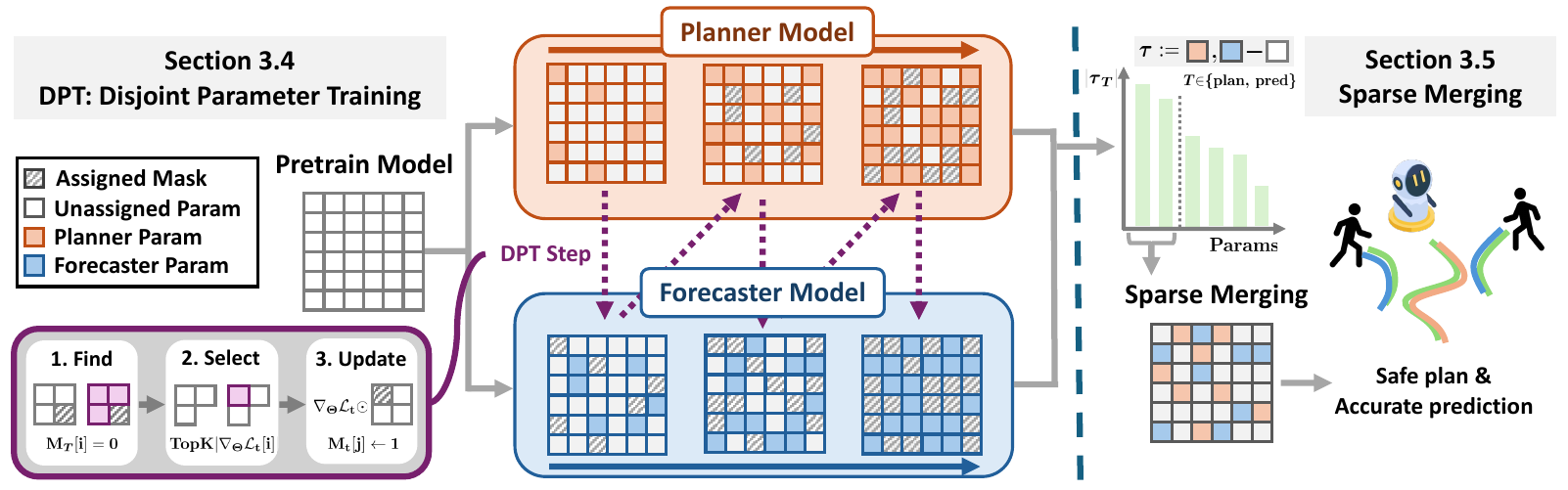}
    \vspace{-5pt}
    \caption{Overview of our proposed framework. In the DPT step, task-specific models for prediction and planning are generated by separating training regions with binary masks to minimize skill conflict. In the Sparse Merging, the two fine-tuned models are combined into a single unified model by utilizing only the most significant task vectors, effectively preserving each task’s skill.}
    \label{fig:overview}
\end{figure*}

\subsection{DPT: Preparing Task-Specialized Material Models}
\label{sec:DPT}
To resolve the observed conflict, we focus on preparing highly disjoint, task-specialized \textit{material} models prior to merging.
We propose \textit{Disjoint Parameter Training} (DPT), a strategy that progressively expands task-critical regions for each task while strictly preventing spatial overlap, as illustrated in Fig.~\ref{fig:overview}.
DPT systematically alternates training between prediction and planning tasks.
At each alternating step, it computes the task-specific loss and gradients across the entire parameter space.
Instead of updating all weights, it selects the top-$K$ unassigned parameters with the largest absolute gradients, permanently allocates them to the current task via a binary mask, and updates only these activated regions.
During the first half of the training process, these task-specific masks incrementally grow.
In the second half, the masks are frozen, allowing the model to consolidate each task's core region without further spatial expansion.
This explicit separation produces fine-tuned models ($f_{\text{plan}}$ and $f_{\text{pred}}$) whose core parameter spaces in the shared encoder are mutually exclusive by construction, drastically reducing interference at merge time.
The complete procedure is summarized in Algorithm~\ref{alg:dpt}.

\begin{algorithm}[t]
\DontPrintSemicolon
\caption{Disjoint Parameter Training (DPT)}\label{alg:dpt}
\KwData{Tasks ${T}={\{\text{plan}, \text{pred}\}}$, total epochs $E$, learning rate $\eta$, mask size $K$, pretrained model $\Theta_0$}
\KwResult{Fine-tuned models $f_{\text{plan}}$, $f_{\text{pred}}$ with disjoint parameter regions}

Initialization: $M_{\text{plan}}, M_{\text{pred}} \gets 0$, $f_{\text{plan}}, f_{\text{pred}} \gets \Theta_0$

\For{$e=1$ \textbf{to} $E/2$}{
\For{$t \in {T}$}{
Compute $\mathcal{L}_t$ and $\nabla_{\Theta} \mathcal{L}_t$ \

\tcp{\small Find unassigned parameters}

$\mathcal{U} = \{\, i \mid M_{\text{plan}}[i] = 0 \;\land\; M_{\text{pred}}[i] = 0 \,\}$ \

\tcp{\small Select Top-$K$ gradient indices}

$\mathcal{I}_{\text{top-}K} \leftarrow \operatorname{TopK}_{i \in \mathcal{U}} |\nabla_{\Theta} \mathcal{L}_t[i]|$  %

\tcp{\small Update mask and model}
$M_t[j] \gets 1$ for $j\in \mathcal{I}_{\text{top-}K}$ \
$f_t \gets f_t - \eta\, \nabla_{\Theta} \mathcal{L}_t \odot M_t$
}
}

\For{$e=E/2$ \textbf{to} $E$}{
\For{$t \in {T}$}{

\tcp{\small Update model with fixed mask}

$f_t \gets f_t - \eta\, \nabla_{\Theta} \mathcal{L}_t \odot M_t$
}
}

Return $f_{\text{plan}}$ and $f_{\text{pred}}$.
\end{algorithm}

\subsection{Sparse Merging of Prepared Models}
\label{sec:sparse_merging}
Although DPT effectively eliminates direct parameter overlap between tasks, we observe that combining the full masked coordinates of the fine-tuned task vectors still degrades joint performance in compact models.
This indicates that even though the fetched parameters do not overlap spatially due to the disjoint masks, local functional interactions still persist: parameter updates in adjacent coordinates or nearby layers can easily couple features.
Consequently, indiscriminately combining full vectors increases cross-task interference, even when their primary supports are largely disjoint.
To address this, we restrict each skill to a narrow, high-signal subset, which yields significantly better composability.
Operationally, we introduce \textbf{Sparse Merging}, which integrates only a highly restrictive fraction of the largest-magnitude coordinates from each task vector utilizing the previously defined mask $M_t^{(K\%)}$.
We define our robust merging operation as:
\begin{equation}
\label{eq:mask}
\Theta_{\mathrm{merged}}=\Theta_{0}+
M_{\text{pred}}^{(\text{K}\%)} \odot \tau_{\text{pred}}+
M_{\text{plan}}^{(\text{K}\%)} \odot \tau_{\text{plan}} .
\end{equation}
By utilizing only the most salient coordinates, Sparse Merging strictly preserves each task’s core skill while actively suppressing detrimental local interactions within the compact shared encoder.
Empirically, this sparse approach consistently outperforms dense merging ($K{=}100\%$) and moderately sparse configurations (e.g., $K{=}10\%$) across all joint metrics, validating that isolating skills into narrow parameter footprints is essential for high-performance multi-task composability on edge devices.

\section{Experiments}
\label{sec:experiments}

\subsection{Experimental Setup}
\noindent\textbf{Datasets \& Backbone.}
We conduct experiments on two datasets that provide visual cues: the JTA~\cite{JTA_fabbri2018learning} and JRDB~\cite{martin2021JRDB} datasets.
JRDB is a real-world dataset collected in diverse indoor and outdoor environments and provides pedestrian trajectories with annotated bounding boxes.
JTA is a large-scale synthetic dataset that provides rich annotations for pedestrian motion.
For both datasets, we follow the standard protocol of predicting the future 12 time steps given the past 9 time steps at 2.5 frames per second (fps).
Additional dataset details are provided in Supplementary Section 1.

The unified backbone is based on Social-Transmotion~\cite{TP_39_ICLR2024_saadatnejad2023social}.
The original architecture, which was designed solely for the prediction task, is extended by adding a planning decoder, enabling the model to perform both prediction and planning jointly.
During pretraining, we employ a game-theoretic loss~\cite{kedia2023GameTheoretic}, one of the popular joint learning paradigms.
Further architectural and learning details of the backbone model are provided in Supplementary Sections 1 and 2.

\noindent\textbf{Baselines.}
Our method aims to achieve both prediction and planning effectively within a single unified model.
Therefore, we compare it against other unified architectures that share the same task objectives (1, 2), as well as multi-task learning approaches (3, 4, 5, 6) that jointly perform related tasks.
(1) \textbf{DIPP}~\cite{huang2023DIPP} is a jointly unified framework of neural network based on a Transformer that uses a differentiable nonlinear optimizer as a motion planner, enabling all operations differentiable.
(2) \textbf{DTPP}~\cite{huang2024DTPP} is also a joint differentiable unified framework that integrates ego conditioned prediction and learnable cost models in a tree structured policy planner.
(3) \textbf{Ensemble}~\cite{arpit2022ensemble} combines the prediction outputs of a planner-finetuned model and a forecaster-finetuned model by averaging their outputs, without modifying any model parameters.
(4) \textbf{Task Arithmetic}~\cite{Merging_Taskarithmetic_ilharco2022editing} merges finetuned models by extracting task vectors from the pretrained model and linearly combines these vectors by weighted averaging.
(5) \textbf{Ties Merging}~\cite{Merging_Tiesmerging_yadav2023ties} selects only the important parameters from each task vector, resolves sign conflicts, and merges the remaining parameters in the direction of sign agreement by averaging them.
(6) \textbf{Localize-and-Stitch}~\cite{Merging_localizeandstich_he2024localize} identifies task-critical parameter regions within fine-tuned models by learning a binary mask and merges only 1\% of a sparse subset into the pretrained model's weights.
(7) \textbf{T-Switch}~\cite{Merging_TSwitch_qi2025less} is a dynamic model merging method that represents task vectors in a binary form and selectively activates task-specific parameter directions through a switching mechanism.

\noindent\textbf{Metrics.}
For planning evaluation, we use core metrics corresponding to the categories of \textit{Effectiveness}, \textit{Safety}, and \textit{Goal Success}, which are essential for safe driving.
ADE (Average Displacement Error) measures average displacement error between the predicted trajectory and the ground-truth trajectory over the full horizon, reflecting \textit{Effectiveness} of trajectory.
Collision Rate quantifies \textit{Safety} by computing the fraction of predicted ego positions that are within a collision threshold of surrounding ground-truth agents; we use a distance threshold of 0.6m.
FDE (Final Displacement Error) measures the displacement error at the final timestep, serving as a proxy for \textit{Goal Success}.
Miss Rate also evaluates \textit{Goal Success} by checking whether the final ego position deviates from the ground truth by more than 0.5m.
For prediction evaluation, we estimate ADE and FDE over neighbor agents by using the same definitions as above.
Detailed definitions of the evaluation metrics are provided in Supplementary Section 3.

\subsection{Experimental Results}

\begin{table*}[t]
\centering
\caption{Applicability of the DPT to different model merging strategies on the JRDB and JTA datasets. The upper rows show task-specific fine-tuned models, while the lower rows show the effect of applying different model merging methods with and without DPT. Sparse Merging is performed with a mask ratio of $K{=}1\%$.}
\label{tab:MainTable2}

\setlength\tabcolsep{6.5pt} %
\renewcommand{\arraystretch}{1.15}
\vspace{-4pt}

\resizebox{\textwidth}{!}{%
\begin{tabular}{l|cccc|cc|cccc|cc}
\toprule
\multirow{4}{*}[-0.5em]{Method}
& \multicolumn{6}{c|}{JRDB}
& \multicolumn{6}{c}{JTA} \\
\cmidrule(lr){2-7}\cmidrule(lr){8-13}

& \multicolumn{4}{c|}{Planning Metric}
& \multicolumn{2}{c|}{Prediction Metric}
& \multicolumn{4}{c|}{Planning Metric}
& \multicolumn{2}{c}{Prediction Metric} \\
\cmidrule(lr){2-5}\cmidrule(lr){6-7}\cmidrule(lr){8-11}\cmidrule(lr){12-13}

& \begin{tabular}[c]{@{}c@{}}Effectiveness\\(ADE$\downarrow$)\end{tabular}
& \begin{tabular}[c]{@{}c@{}}Safety\\(Col.\ rate$\downarrow$)\end{tabular}
& \begin{tabular}[c]{@{}c@{}}Goal Success\\(FDE$\downarrow$)\end{tabular}
& \begin{tabular}[c]{@{}c@{}}Goal Success\\(Miss rate$\downarrow$)\end{tabular}
& ADE$\downarrow$ & FDE$\downarrow$
& \begin{tabular}[c]{@{}c@{}}Effectiveness\\(ADE$\downarrow$)\end{tabular}
& \begin{tabular}[c]{@{}c@{}}Safety\\(Col.\ rate$\downarrow$)\end{tabular}
& \begin{tabular}[c]{@{}c@{}}Goal Success\\(FDE$\downarrow$)\end{tabular}
& \begin{tabular}[c]{@{}c@{}}Goal Success\\(Miss rate$\downarrow$)\end{tabular}
& ADE$\downarrow$ & FDE$\downarrow$ \\
\midrule

Plan Finetune
& 0.3965 & 0.0078 & 0.7371 & 0.3700 & 1.9825 & 2.4087
& 1.3273 & 0.0056 & 2.6118 & 0.9144 & 8.5329 & 9.8826 \\
\rowcolor[HTML]{EFEFEF}
Plan Finetune + DPT
& 0.3699 & 0.0073 & 0.7244 & 0.3600 & 0.7368 & 1.1702
& 1.0008 & 0.0055 & 2.0300 & 0.8877 & 1.2079 & 2.3390 \\

\addlinespace[2pt]

Pred Finetune
& 0.8889 & 0.0195 & 1.2140 & 0.9615 & 0.4967 & 0.9144
& 5.1324 & 0.0052 & 8.9957 & 0.9960 & 1.4394 & 2.7745 \\
\rowcolor[HTML]{EFEFEF}
Pred Finetune + DPT
& 0.8469 & 0.0200 & 1.1244 & 0.9189 & 0.5250 & 0.9682
& 1.1139 & 0.0058 & 2.2594 & 0.9612 & 1.1834 & 2.3184 \\
\midrule

Task Arithmetic~\cite{Merging_Taskarithmetic_ilharco2022editing}
& 0.9387 & 0.0189 & 1.1798 & 0.8887 & 0.9904 & 1.3870
& 3.4726 & 0.0058 & 6.1710 & 0.9973 & 4.1518 & 5.7700 \\
\rowcolor[HTML]{DAE8FC}
Task Arithmetic + DPT
& \textbf{0.5865} & \textbf{0.0145} & \textbf{0.8859} & \textbf{0.4283} & \textbf{0.5849} & \textbf{1.0308}
& \textbf{1.0373} & \textbf{0.0056} & \textbf{2.0936} & \textbf{0.9278} & \textbf{1.1406} & \textbf{2.2367} \\

\addlinespace[2pt]

Ties Merging~\cite{Merging_Tiesmerging_yadav2023ties}
& 0.8759 & 0.0167 & 1.1503 & 0.9019 & 0.9463 & 1.3412
& 2.3432 & 0.0067 & 3.2703 & 0.9786 & 3.6955 & 4.6017 \\
\rowcolor[HTML]{DAE8FC}
Ties Merging + DPT
& \textbf{0.4193} & \textbf{0.0100} & \textbf{0.7346} & \textbf{0.3731} & \textbf{0.5917} & \textbf{1.0442}
& \textbf{1.0354} & \textbf{0.0055} & \textbf{2.1154} & \textbf{0.9291} & \textbf{1.1371} & \textbf{2.2277} \\
\midrule

Sparse Merging
& 0.9246 & 0.0185 & 1.1694 & 0.9610 & 0.6587 & 1.0873
& 1.9675 & 0.0075 & 3.1188 & 0.9519 & 1.7600 & 2.7035 \\
\rowcolor[HTML]{DAE8FC}
Sparse Merging + DPT
& \textbf{0.4044} & \textbf{0.0091} & \textbf{0.7458} & \textbf{0.3706} & \textbf{0.5952} & \textbf{1.0352}
& \textbf{1.0322} & \textbf{0.0055} & \textbf{2.1134} & \textbf{0.9305} & \textbf{1.1372} & \textbf{2.2206} \\
\bottomrule
\end{tabular}%
}

\end{table*}

\begin{table}[t]
\centering
\caption{Performance comparison according to task-specific parameter allocation in DPT on the JRDB dataset with Sparse Merging. Sparse Merging is performed with a mask ratio of $K{=}1\%$. A higher DPT allocation ratio leads to lower performance in the corresponding task.}
\label{tab:MainTable3}

\setlength\tabcolsep{6.0pt}  %
\renewcommand{\arraystretch}{1.15}
\vspace{-4pt}

\resizebox{0.98\columnwidth}{!}{%
\begin{tabular}{cc|cccc|cc}
\toprule
\multicolumn{2}{c|}{DPT allocation ratio} &
\multicolumn{4}{c|}{Planning Metric} &
\multicolumn{2}{c}{Prediction Metric} \\
\cmidrule(lr){1-2}\cmidrule(lr){3-6}\cmidrule(lr){7-8}
Plan (\%) & Pred (\%) &
\begin{tabular}[c]{@{}c@{}}Effectiveness\\(ADE $\downarrow$)\end{tabular} &
\begin{tabular}[c]{@{}c@{}}Safety\\(Col.\ rate $\downarrow$)\end{tabular} &
\begin{tabular}[c]{@{}c@{}}Goal Success\\(FDE $\downarrow$)\end{tabular} &
\begin{tabular}[c]{@{}c@{}}Goal Success\\(Miss rate $\downarrow$)\end{tabular} &
ADE $\downarrow$ & FDE $\downarrow$ \\
\midrule

90 & 10 &
\cellcolor{blue!10}{0.4899} &
\cellcolor{blue!10}{0.0156} &
\cellcolor{blue!10}{0.7748} &
\cellcolor{blue!10}{0.3802} &
\cellcolor{red!39}{0.5777} &
\cellcolor{red!42}{1.0139} \\

75 & 25 &
\cellcolor{blue!29}{0.4352} &
\cellcolor{blue!27}{0.0121} &
\cellcolor{blue!30}{0.7515} &
\cellcolor{blue!29}{0.3733} &
\cellcolor{red!45}{\textbf{0.5724}} &
\cellcolor{red!45}{\textbf{1.0107}} \\

50 & 50 &
\cellcolor{blue!40}{0.4044} &
\cellcolor{blue!41}{0.0091} &
\cellcolor{blue!35}{0.7458} &
\cellcolor{blue!37}{0.3706} &
\cellcolor{red!20}{0.5952} &
\cellcolor{red!19}{1.0352} \\

25 & 75 &
\cellcolor{blue!42}{0.3983} &
\cellcolor{blue!45}{\textbf{0.0083}} &
\cellcolor{blue!36}{0.7449} &
\cellcolor{blue!35}{0.3711} &
\cellcolor{red!18}{0.5972} &
\cellcolor{red!16}{1.0375} \\

10 & 90 &
\cellcolor{blue!45}{\textbf{0.3898}} &
\cellcolor{blue!45}{\textbf{0.0083}} &
\cellcolor{blue!45}{\textbf{0.7349}} &
\cellcolor{blue!45}{\textbf{0.3676}} &
\cellcolor{red!10}{0.6046} &
\cellcolor{red!10}{1.0436} \\
\bottomrule
\end{tabular}%
}

\vspace{-10pt}
\end{table}

\noindent\textbf{Applicability of DPT to Other Merging Methods.}
Existing merging methods mainly focus on combining task vectors that are already fine-tuned, whereas DPT prevents Skill Conflict during task-specific training, i.e., in the process of forming task vectors for each task.
Therefore, the proposed DPT framework can be effectively applied not only to our Sparse Merging but also to other model merging methods.
As shown in Tab.~\ref{tab:MainTable2}, we conducted experiments comparing the performance before and after applying DPT to existing model merging methods, including Task Arithmetic~\cite{Merging_Taskarithmetic_ilharco2022editing}, Ties Merging~\cite{Merging_Tiesmerging_yadav2023ties}, and our Sparse Merging. Sparse merging is performed using the mask ratio of K=1\%.
\textit{Plan Finetune} and \textit{Pred Finetune} represent the performance of models fine-tuned on each respective task, and the fact that their performance remains largely unchanged even with DPT indicates that our method does not degrade the inherent capability of the model.
Moreover, when merging is performed using DPT-based fine-tuned models, all merging methods exhibit overall performance improvements.
This trend is consistently observed on both the JRDB and JTA datasets.
These results indicate that DPT is broadly applicable regardless of the merging strategy or dataset domain.

\noindent\textbf{Task-Specific Parameter Allocation of DPT.}
DPT performance is subject to variation based on the extent of task-specific parameter allocation.
As shown in Tab.~\ref{tab:MainTable3}, we conducted experiments under the Sparse Merging setting, where the mask ratio is K=1\%.
We analyzed how the DPT allocation ratio, which defines the maximum range of learnable parameters during task-specific fine-tuning, affects performance.
The results show that as the DPT allocation ratio increases for a particular task, its performance tends to decrease.
This indicates that when the learnable parameter space is large, the task’s skill becomes distributed across a wider range of parameters, requiring the use of more task vectors beyond the top 1\%.
This observation supports the finding discussed in Sec.~\ref{sec:skill_conflict}, which suggests that in unified models, improving single-task performance requires leveraging a larger number of task vectors.
In contrast, when the learnable region is smaller, DPT still achieves sufficient task performance through the Sparse Merging strategy, as it learns compactly from a localized parameter region.
These results demonstrate that DPT effectively captures each task’s skill in a compact manner, enabling it to alleviate \textit{Skill Conflict} within the unified model.

\noindent\textbf{Effectiveness of Sparse Merging in DPT.}
Even when applying DPT to completely separate the trainable parameter regions, interference can still occur during merging if the corresponding features are adjacent.
Therefore, we conduct a performance comparison experiment of \textit{Sparse Merging} according to the \textit{mask ratio} $K$, as shown in Tab.~\ref{tab:MainTable4}.
In this experiment, the maximum parameter allocation ratio for DPT is set to 50:50 by default, and evaluations are performed on the JRDB dataset.
When the sparsity is high ($K{=}2$), the model achieves the best overall performance across all metrics, while performance gradually degraded as the task utilization ratio increased.
This demonstrates that our Sparse Merging effectively minimizes interference between adjacent parameter regions during merging.

\begin{table}[t]
\centering

\begin{minipage}[t]{0.54\linewidth} %
\centering
\captionsetup{type=table}

\caption{Performance comparison by \textit{mask ratio} $K$ under Sparse Merging. The experiment uses a 50:50 DPT allocation on JRDB.}
\label{tab:MainTable4}

\small %
\setlength{\tabcolsep}{4.2pt}       %
\renewcommand{\arraystretch}{1.35}  %
\vspace{-4pt}

\resizebox{0.98\linewidth}{!}{%
\begin{tabular}{c|cccc|cc}
\toprule
\multirow{2}{*}{\raisebox{-1ex}{$K$}} &
\multicolumn{4}{c|}{Planning Metric} &
\multicolumn{2}{c}{Prediction Metric} \\
\cmidrule(lr){2-5}\cmidrule(lr){6-7}
& \begin{tabular}[c]{@{}c@{}}Effectiveness\\(ADE $\downarrow$)\end{tabular}
& \begin{tabular}[c]{@{}c@{}}Safety\\(Col.\ rate $\downarrow$)\end{tabular}
& \begin{tabular}[c]{@{}c@{}}Goal Success\\(FDE $\downarrow$)\end{tabular}
& \begin{tabular}[c]{@{}c@{}}Goal Success\\(Miss rate $\downarrow$)\end{tabular}
& ADE $\downarrow$
& FDE $\downarrow$ \\
\midrule
$1$  & \cellcolor{blue!44}{0.4044} & \cellcolor{blue!45}{\textbf{0.0091}} & \cellcolor{blue!10}{0.7458} & \cellcolor{blue!39}{0.3706} & \cellcolor{red!10}{0.5952} & \cellcolor{red!26}{1.0352} \\
$2$  & \cellcolor{blue!45}{\textbf{0.4041}} & \cellcolor{blue!28}{0.0096} & \cellcolor{blue!42}{0.7320} & \cellcolor{blue!45}{\textbf{0.3697}} & \cellcolor{red!45}{\textbf{0.5753}} & \cellcolor{red!45}{\textbf{1.0174}} \\
$5$  & \cellcolor{blue!24}{0.4156} & \cellcolor{blue!14}{0.0100} & \cellcolor{blue!45}{\textbf{0.7306}} & \cellcolor{blue!29}{0.3722} & \cellcolor{red!24}{0.5871} & \cellcolor{red!30}{1.0322} \\
$10$ & \cellcolor{blue!17}{0.4193} & \cellcolor{blue!14}{0.0100} & \cellcolor{blue!36}{0.7346} & \cellcolor{blue!23}{0.3730} & \cellcolor{red!16}{0.5917} & \cellcolor{red!17}{1.0442} \\
$20$ & \cellcolor{blue!13}{0.4213} & \cellcolor{blue!10}{0.0101} & \cellcolor{blue!26}{0.7389} & \cellcolor{blue!10}{0.3750} & \cellcolor{red!16}{0.5920} & \cellcolor{red!10}{1.0507} \\
$40$ & \cellcolor{blue!10}{0.4229} & \cellcolor{blue!10}{0.0101} & \cellcolor{blue!21}{0.7410} & \cellcolor{blue!12}{0.3747} & \cellcolor{red!20}{0.5895} & \cellcolor{red!10}{1.0505} \\
\bottomrule
\end{tabular}%
}
\end{minipage}
\hfill
\begin{minipage}[t]{0.43\linewidth} %
\centering
\captionsetup{type=figure}
\includegraphics[width=\linewidth]{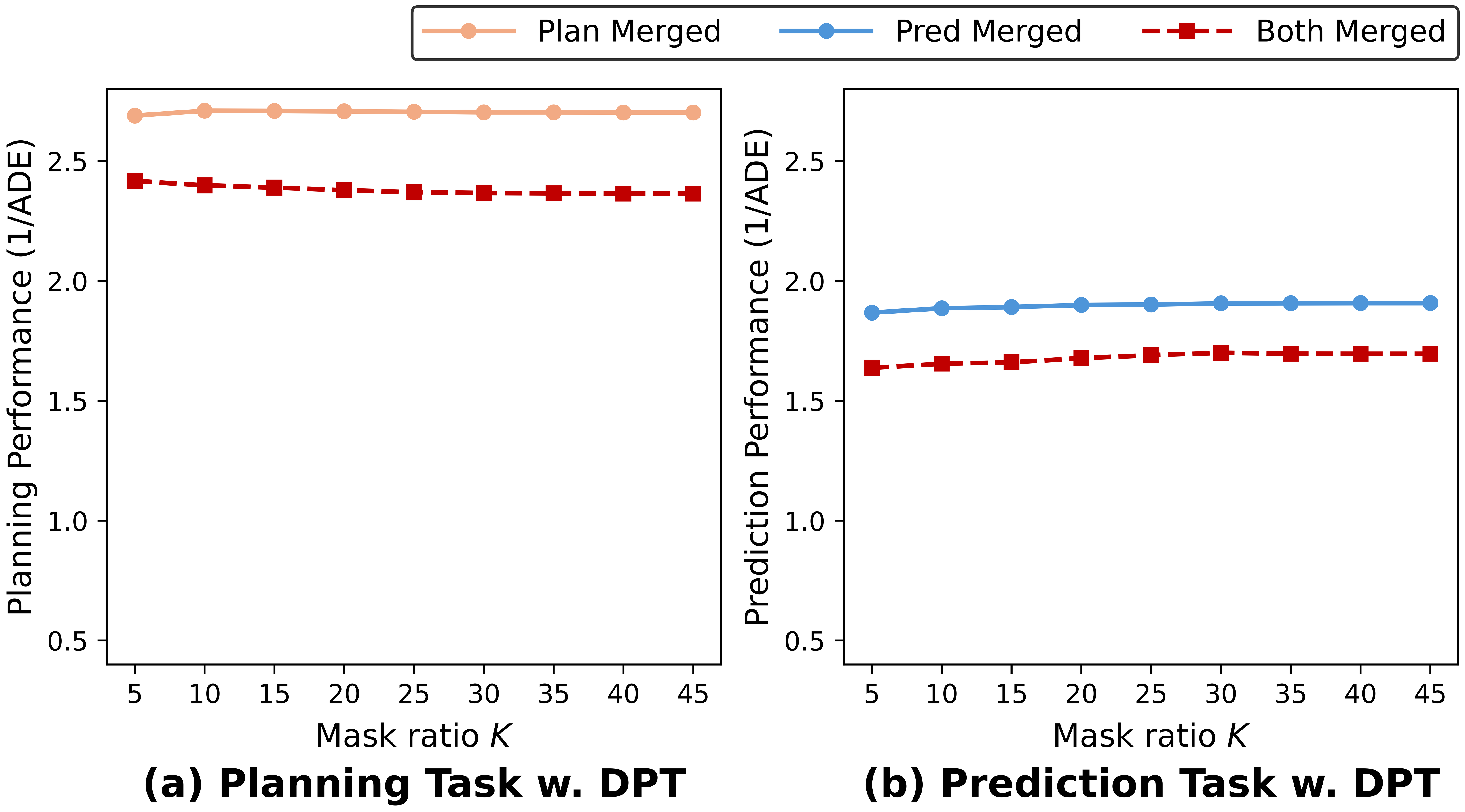}
\vspace{-6pt}
\caption{Performance of (a) Planning and (b) Prediction tasks trained with DPT depends on \textit{mask ratio} $K$.}
\label{fig:fig_5_DPT_perf_graph}
\end{minipage}

\end{table}

\noindent\textbf{Mitigating Skill Conflict through DPT.}
To verify whether DPT effectively resolves the \textit{Skill Conflict} observed in Fig.~\ref{fig:fig_3_ordinary_perf_graph}, we conduct the same experiments with DPT applied.
As shown in Fig.~\ref{fig:fig_5_DPT_perf_graph}, even when a large number of parameters are allocated to each task, DPT maintains high performance for both tasks, unlike the results in Fig.~\ref{fig:fig_3_ordinary_perf_graph}.
Moreover, even with a smaller parameter allocation, DPT quickly achieves high single-task performance.
This is because DPT begins training within a narrow parameter region and gradually expands the learnable region, allowing the model to encode each task’s skill within a compact area.
These results show that DPT effectively mitigates \textit{Skill Conflict} and enables each task's capability to be represented efficiently using a small set of parameters.

\noindent\textbf{Comparison to Baselines.}
Tab.~\ref{tab:MainTable1} compares the proposed DPT and Sparse Merging framework with various baselines on the JRDB dataset.
All experiments are conducted with a mask ratio of $K{=}1\%$ in Sparse Merging.
Compared with other unified model architectures such as DIPP~\cite{huang2023DIPP} and DTPP~\cite{huang2024DTPP}, our method achieves superior performance across planning and prediction metrics.
This suggests that the task conflict between planning and prediction also exists in other unified model architectures, leading to suboptimal performance for both tasks.
Among the merging baselines, the Ensemble method, which averages the inference results of each fine-tuned model, performs notably better than other merging strategies.
This observation indicates that the conflict between planning and prediction primarily arises at the parameter level rather than only at the output level, supporting the existence of the previously identified \textit{Skill Conflict}.
Overall, the proposed DPT combined with Sparse Merging achieves the highest performance across all planning and prediction metrics on the JRDB dataset, demonstrating the effectiveness of our framework.

\begin{table}[t]
\centering
\caption{Quantitative comparison between the proposed DPT with Sparse Merging framework and the baselines on the JRDB dataset. Sparse Merging is performed using a mask ratio of $K{=}1\%$. We also compare the results obtained by applying joint reasoning within the existing framework. The comparison baselines are categorized into unified models with different architectures and methods applicable in parallel to our unified model, including ensemble and merging-based approaches. (SM: Sparse Merging; JR: Joint Reasoning)}
\label{tab:MainTable1}

\setlength\tabcolsep{6.0pt}
\renewcommand{\arraystretch}{1.15}
\vspace{-4pt}

\resizebox{0.98\linewidth}{!}{%
\begin{tabular}{l|cccc|cc}
\toprule
\multirow{3}{*}[-0.4em]{Method}
& \multicolumn{4}{c|}{Planning Metric}
& \multicolumn{2}{c}{Prediction Metric} \\
\cmidrule(lr){2-5}\cmidrule(lr){6-7}
& \begin{tabular}[c]{@{}c@{}}Effectiveness\\(ADE $\downarrow$)\end{tabular}
& \begin{tabular}[c]{@{}c@{}}Safety\\(Col.\ rate $\downarrow$)\end{tabular}
& \begin{tabular}[c]{@{}c@{}}Goal Success\\(FDE $\downarrow$)\end{tabular}
& \begin{tabular}[c]{@{}c@{}}Goal Success\\(Miss rate $\downarrow$)\end{tabular}
& ADE $\downarrow$
& FDE $\downarrow$ \\
\midrule

DIPP~\cite{huang2023DIPP}
& 0.8048 & 0.0167 & 1.3542 & 0.6940 & 0.8834 & 1.6406 \\

DTPP~\cite{huang2024DTPP}
& 0.6232 & 0.0145 & 1.1338 & 0.6931 & 0.7686 & 1.4253 \\

\midrule

Ensemble~\cite{arpit2022ensemble}
& 0.5807 & 0.0146 & 0.9000 & 0.4786 & 1.1313 & 1.5258 \\

Task Arithmetic~\cite{Merging_Taskarithmetic_ilharco2022editing}
& 0.9387 & 0.0189 & 1.1798 & 0.8887 & 0.9904 & 1.3870 \\

Ties Merging~\cite{Merging_Tiesmerging_yadav2023ties}
& 0.8759 & 0.0167 & 1.1503 & 0.9019 & 0.9463 & 1.3412 \\

Localize-and-stitch~\cite{Merging_localizeandstich_he2024localize}
& 0.8838 & 0.0176 & 1.1393 & 0.9487 & 0.8104 & 1.2670 \\

T-Switch~\cite{Merging_TSwitch_qi2025less}
& 0.8047 & 0.0199 & 1.1009 & 0.8784 & 0.8091 & 1.2112 \\

\midrule

\rowcolor[HTML]{EEF5FF}
DPT + SM
& \underline{0.4044} & \underline{0.0091} & \underline{0.7458} & \underline{0.3706} & \underline{0.5952} & \underline{1.0352} \\

\rowcolor[HTML]{DAE8FC}
DPT + SM + JR
& \textbf{0.3640} & \textbf{0.0074} & \textbf{0.7105} & \textbf{0.3589} & \textbf{0.5097} & \textbf{0.9396} \\
\bottomrule
\end{tabular}%
}
\vspace{-10pt}

\end{table}

\noindent\textbf{Effect of Adding Joint Reasoning.}
The DPT and Sparse Merging strategies reduce inter-task conflicts by separating the parameter regions used by each task. However, this separation may weaken the inherent advantage of unified models—namely, joint reasoning across tasks. To preserve the task-specific characteristics while still enabling joint reasoning, we perform an additional fine-tuning stage using a joint task loss on the parameter regions that remain inactive after DPT and Sparse Merging.
As shown in Tab.~\ref{tab:MainTable1}, models that incorporate this joint reasoning stage achieve better performance than those that only separate task-specific parameters. This highlights the importance of joint reasoning in unified models and demonstrates that, by isolating the core task-specific parameter regions while allowing the remaining parameters to support joint reasoning, our approach is more effective than conventional unified training schemes (DIPP~\cite{huang2023DIPP}, DTPP~\cite{huang2024DTPP}, Pretrained Model).

\begin{figure*}[t]
    \centering
    \includegraphics[width=\textwidth]{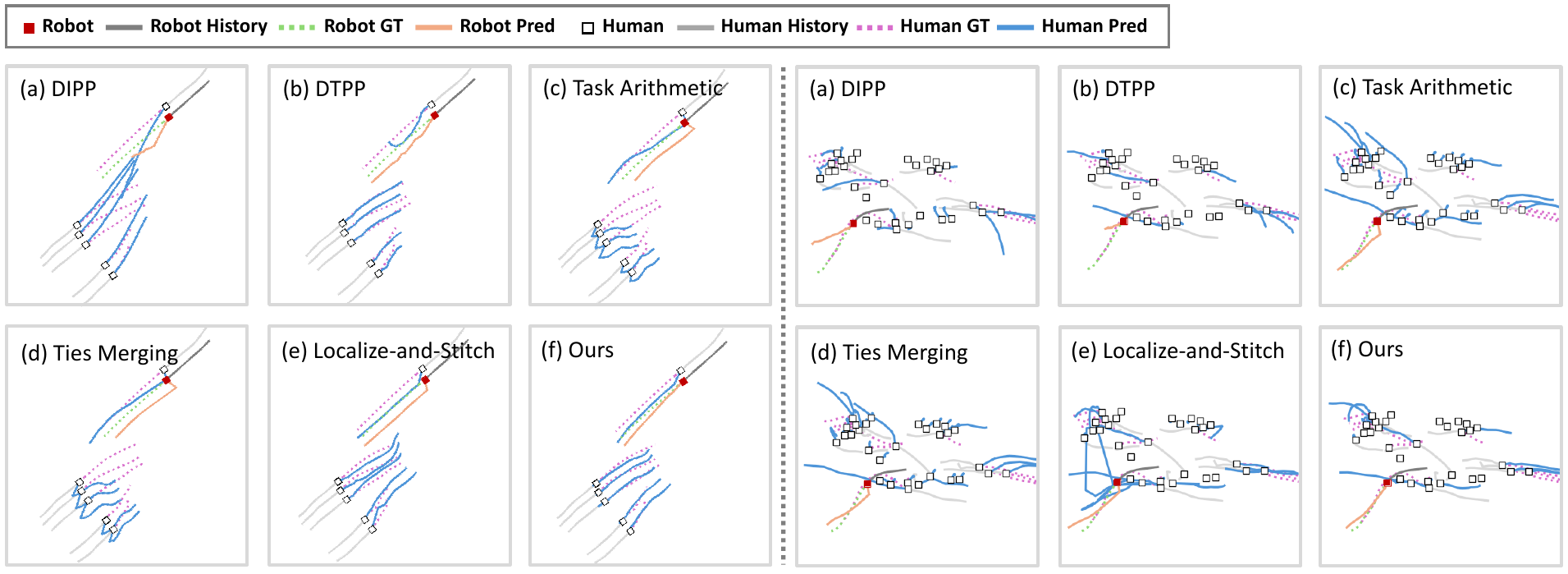}
    \caption{Qualitative Results on the JRDB dataset. (a)–(b) show unified models with different architectures, while (c)–(e) present results of applying various baseline model merging methods on our backbone-based unified model. These are compared with (f) Ours, which employs DPT and Sparse Merging. In both examples, our method accurately predicts surrounding agents and produces safer plans compared to the baselines.}
    \label{fig:fig_6_Qualitative_results}
    \vspace{-10pt}
\end{figure*}

\noindent\textbf{Qualitative Results.}
Fig.~\ref{fig:fig_6_Qualitative_results} compares qualitative results on the JRDB dataset.
(a) and (b) show the results of other unified model architectures, DIPP and DTPP, while (c)–(e) present the results of applying different baseline model-merging methods on our unified-model backbone.
These are compared with (f) Ours, which employs the proposed DPT and Sparse Merging approach.
In the left sample of Fig.~\ref{fig:fig_6_Qualitative_results}, both (a) DIPP and (b) DTPP exhibit generally poor prediction and planning performance for nearby agents, indicating that conventional unified models fail to effectively handle skill conflict between tasks.
Meanwhile, in (e) Localize-and-Stitch, only a subset of key parameter regions from the fine-tuned models is added to the pretrained model. As a result, the behavior of the pretrained model—characterized by aggressive prediction and safe planning learned in a game-theoretic manner—remains distinctly reflected.
Other model-merging baselines show similar tendencies, suggesting that existing merging methods fail to fully exploit the fine-tuned task skills due to skill conflict.
In contrast, our approach—performing task-wise disjoint fine-tuning via DPT and applying Sparse Merging to minimize interference between adjacent features—achieves both accurate prediction and safe planning.
This trend consistently appears even in dense scenes, as shown in the right sample of Fig.~\ref{fig:fig_6_Qualitative_results}, demonstrating that our method effectively mitigates skill conflict in unified models and achieves strong performance across both tasks.

Additional analyses are provided in the supplementary material, including additional qualitative results in dense scenes (Suppl. Sec. 5), extended empirical studies of Skill Conflict across different architectures, model capacities, DPT settings and random seeds, controlled overlap, sub-task decomposition, mask evolution, and layer-wise conflict analysis (Suppl. Sec. 6), additional baseline comparisons with T-Switch, MTL baselines, and full-JTA results (Suppl. Sec. 7), training cost and inference efficiency analysis (Suppl. Sec. 8), and an extension to end-to-end autonomous driving with closed-loop and open-loop evaluation in the vehicle domain (Suppl. Sec. 9).

\section{Conclusion}
\label{sec:conclusion}

In this paper, we identify the Skill Conflict that arises within the shared encoder of multiple unified models through an empirical study, and propose two sequential methods---Disjoint Parameter Training (DPT) and Sparse Merging---to address it. DPT mitigates task interference by separating the learnable parameter regions of the prediction and planning tasks when constructing task-specialized material models. Sparse Merging performs merging using only the core task vectors to mitigate residual conflicts that may still occur in adjacent features, even when the main parameter regions are separated. This framework outperforms existing baselines and demonstrates the applicability of DPT to various merging methods. Overall, our results suggest that explicitly controlling parameter-level task interference is important for achieving accurate prediction and safe planning within compact unified models.

\section*{Acknowledgements}
This work was supported by the National Research Foundation of Korea (NRF) grants funded by the Korea government (MSIT) (No. RS-2026-25494733 and No. RS-2025-22802992), the Institute of Information \& Communications Technology Planning \& Evaluation (IITP) grants funded by the Korea government (MSIT) (No. RS-2025-25442149, LG AI STAR Talent Development Program for Leading Large-Scale Generative AI Models in the Physical AI Domain, and No. RS-2025-02219277, AI Star Fellowship Support (DGIST)), the Basic Science Research Program through the National Research Foundation of Korea (NRF) funded by the Ministry of Education (No. RS-2025-25420118), and the InnoCORE program of the Ministry of Science and ICT (No. N10260003 and No. 26-InnoCORE-01).

\vspace{\baselineskip}
\noindent Research supported by the NVIDIA Academic Grant Program.

\clearpage
\title{Unified Prediction and Planning via Conflict-Aware Disjoint Parameter Training \\ \vspace{0.5cm} \Large {Supplementary Material}}
\titlerunning{Conflict-Aware Disjoint Parameter Training}
\author{Taewon Seo\inst{1}$^\star$\orcidlink{0009-0007-1022-3354} \and
Seonae Jeon\inst{1}$^\star$\orcidlink{0009-0006-4452-3542} \and
Giwon Lee\inst{2}$^\star$\orcidlink{0009-0006-5472-2078} \and
Kuk-Jin Yoon \inst{2}$^\dagger$\orcidlink{0000-0002-1634-2756} \and
Daehee Park \inst{1}$^\dagger$\orcidlink{0000-0002-3961-6932}}
\authorrunning{Seo et al.}
\institute{
DGIST, Republic of Korea\\
\email{\{taewonseo,seonae,dhpark\}@dgist.ac.kr}
\and
KAIST, Republic of Korea\\
\email{\{dlrldnjs,kjyoon\}@kaist.ac.kr}
}
\maketitle

\renewcommand{\theHsection}{supp.\arabic{section}}
\renewcommand{\theHsubsection}{supp.\arabic{section}.\arabic{subsection}}
\renewcommand{\theHsubsubsection}{supp.\arabic{section}.\arabic{subsection}.\arabic{subsubsection}}
\renewcommand{\theHequation}{supp.eq.\arabic{equation}}
\renewcommand{\theHfigure}{supp.fig.\arabic{figure}}
\renewcommand{\theHtable}{supp.tab.\arabic{table}}

\begingroup
\renewcommand\thefootnote{}
\footnotetext{
$^\star$ Equal contribution. \quad
$^\dagger$ Corresponding authors.
}
\endgroup

In this supplementary material, we provide detailed descriptions of the experimental settings used in our study, along with additional empirical analyses and experimental results. We also include, in Section 9, extended experimental results that demonstrate the applicability of our method to an end-to-end autonomous driving (E2E-AD) model in the vehicle domain.

\begin{itemize}
\item Sec.~\ref{sup:Datasets_Details} provides comprehensive details regarding the JRDB and JTA datasets.
\item Sec.~\ref{sup:Backbone} describes the unified backbone architecture and the training process designed to encourage game-theoretic interaction between planning and prediction.
\item Sec.~\ref{sup:evaluation_metrics} defines the evaluation metrics used to assess performance.
\item Sec.~\ref{sup:Implemenrtation_details} outlines the specific implementation details, such as the hardware specifications, optimizers, and hyperparameters utilized for each dataset.
\item Sec.~\ref{sup:Qualitative_results} presents additional qualitative results that demonstrate the effectiveness and robustness of the DPT + Sparse Merging approach in dense scenes.
\item Sec.~\ref{sup:Empirical_Study} provides an empirical study verifying that the skill conflict phenomenon is a pervasive issue.
\item Sec.~\ref{sup:Additional_Baseline_Comparisons} explores the compatibility of DPT with other baselines.
\item Sec.~\ref{sup:Training_Cost_Comparison} compares the training efficiency and computational costs of our framework.
\item Sec.~\ref{sup:Adaptability_on_E2EAD} includes extended experimental results that demonstrate the applicability of our method to an end-to-end autonomous driving (E2E-AD) model in the vehicle domain.
\item Sec.~\ref{sub:future_direction} discusses future directions for extending DPT to broader multi-task settings and hyperparameter design.
\end{itemize}

\section{Dataset Details}
\label{sup:Datasets_Details}

Our experiments are conducted on two datasets: JRDB and JTA.
JRDB is a real-world dataset collected in both indoor and outdoor environments.
It provides pedestrian trajectories and annotated bounding boxes with high variability across diverse scenes.
In our experiments, we utilize 2D trajectories and 2D bounding boxes.
For the JRDB dataset, we use 19,843 samples for training, 496 for validation, and 6,341 for testing.
The JTA dataset is a large-scale synthetic dataset consisting of 256 training sequences, 128 validation sequences, and 128 test sequences, with approximately 10 million annotated 3D keypoints.
We utilize 2D trajectories, 2D/3D bounding boxes, and 2D/3D poses, which enables a comprehensive analysis of pedestrian motion in complex scenarios.
Since the JTA dataset is significantly larger than JRDB, we use only 15\% of the full dataset to maintain a comparable experimental setup between the two datasets and to minimize any bias that may arise from large differences in dataset size.
Within this 15\% subset, 13,221 samples are used for training, 530 for validation, and 748 for testing.
For reference, the full JTA dataset contains 88,719 samples for training, 3,565 for validation, and 5,021 for testing.

\section{Backbone Model}
\label{sup:Backbone}

\begin{figure}[t]
    \centering
    \includegraphics[width=0.65\linewidth]{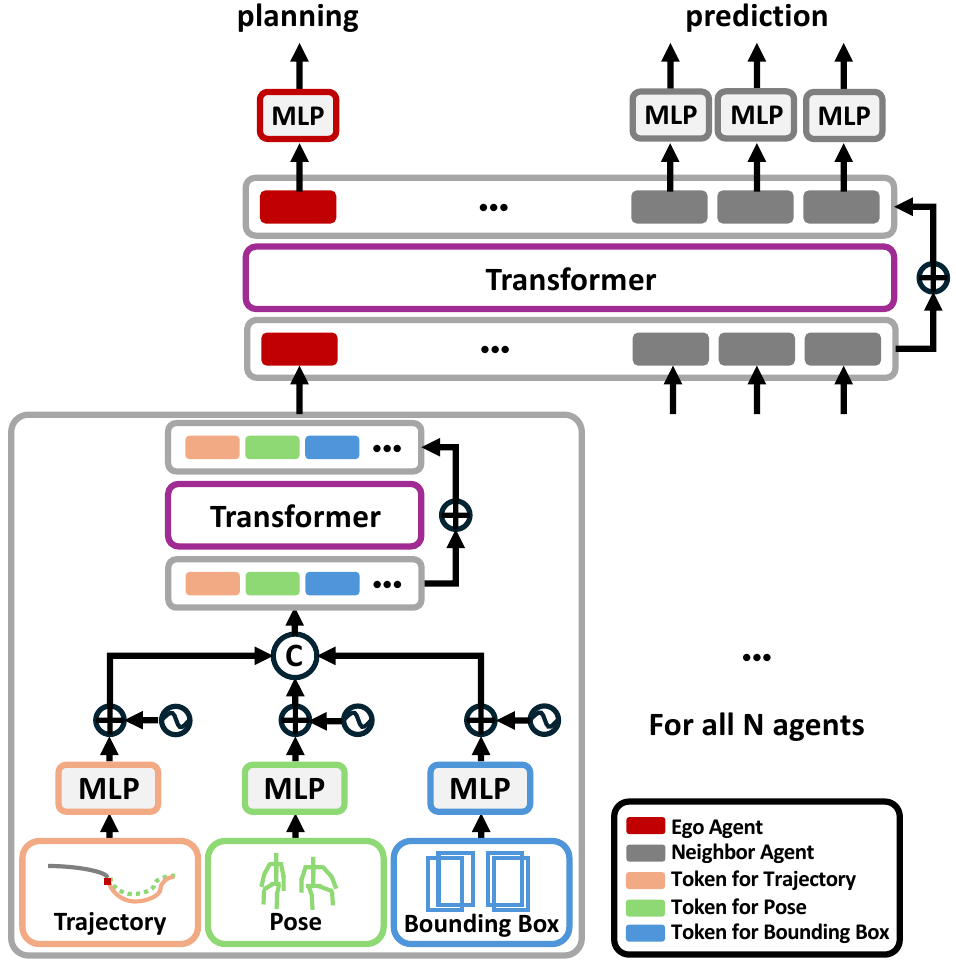}
    \caption{Overview of the unified backbone architecture. Our model extends the Social-Transmotion backbone by preserving its multi-modal tokenization—trajectory, pose, and bounding box embeddings—and processing all agent-wise tokens through a shared Transformer encoder. There are two task-specific decoders: a planning decoder that produces the ego agent's future trajectory, and a prediction decoder that forecasts trajectories of all neighbor agents.
    }
    \label{fig:sup_fig_backbone_architecture}
\end{figure}

Our unified model is based on Social-Transmotion~\cite{supp-TP_39_ICLR2024_saadatnejad2023social}, and its overall architecture is illustrated in Fig.~\ref{fig:sup_fig_backbone_architecture}.
The original Social-Transmotion architecture performs prediction only; however, in this study, we extend it to a unified architecture that handles both planning and prediction by adding a planning decoder.
The training process consists of two stages: (1) joint pretraining, where the two tasks are learned together, and (2) task-specific fine-tuning, which improves the performance of each task.
During joint pretraining, we alternately optimize the parameters of the planner and the forecaster. This alternating update encourages the game-theoretic interaction~\cite{supp-huang2023gameformer} between the two modules.
The loss function for the planner step is defined as:
\begin{equation}
    \mathcal{L}_\text{plan} = \mathcal{L}^{\text{ego}}_{\text{ADE}} + \lambda_{\text{col}} \cdot \Delta\mathcal{C}_{\text{plan}}.
\end{equation}
where $\mathcal{L}^{\text{ego}}_{\text{ADE}}$ is the Average Displacement Error between the ground-truth ego trajectory $Y_{\text{plan}}$ and the predicted ego trajectory $\hat{Y}_{\text{plan}}$.
$\Delta\mathcal{C}_{\text{plan}}$ denotes the relative collision cost, computed based on the forecaster's predicted neighbor agents' trajectories:
\begin{equation}
    \Delta\mathcal{C}_{\text{plan}} = \mathcal{C}(Y_{\text{plan}}, Y_{\text{pred}}) - \mathcal{C}(\hat{Y}_{\text{plan}}, Y_{\text{pred}}).
\end{equation}
The loss function for the forecaster step is:
\begin{equation}
    \mathcal{L}_\text{pred} = \mathcal{L}^{\text{neigh}}_{\text{ADE}} + \lambda_{\text{col}} \cdot \Delta\mathcal{C}_{\text{pred}}.
\end{equation}
where $\mathcal{L}^{\text{neigh}}_{\text{ADE}}$ is the ADE between the ground-truth neighbor trajectories $Y_{\text{pred}}$ and the predicted neighbor trajectories $\hat{Y}_{\text{pred}}$.
$\Delta\mathcal{C}_{\text{pred}}$ is computed based on the planner’s predicted ego trajectory:
\begin{equation}
    \Delta\mathcal{C}_{\text{pred}} = \mathcal{C}(Y_{\text{plan}}, Y_{\text{pred}}) - \mathcal{C}(Y_{\text{plan}}, \hat{Y}_{\text{pred}}).
\end{equation}
After pretraining, we perform task-specific fine-tuning.
During planner fine-tuning, the forecaster’s predictions are fixed, and the planner is optimized to generate accurate and safe trajectories.
\begin{equation}
    \mathcal{L}_{\text{plan-ft}} = \mathcal{L}_{\text{ADE}}^{\text{ego}} + \lambda_{\text{col}} \cdot \left( \mathcal{C}(Y_{\text{plan}}, Y_{\text{pred}}^{\text{freeze}}) - \mathcal{C}(\hat{Y}_{\text{plan}}, Y_{\text{pred}}^{\text{freeze}}) \right)
\end{equation}
For forecaster fine-tuning, the objective consists solely of the pure ADE loss, excluding the collision cost term:
\begin{equation}
    \mathcal{L}_{\text{pred-ft}} = \mathcal{L}_{\text{ADE}}^{\text{neigh}}
\end{equation}
To quantify the potential collision risk between the ego agent and surrounding neighbor agents, we employ the following collision cost function.
At each time step $t$, we compute the Euclidean distance between the ego position $Y_0^{(t)}$ and the position of neighbor agent $n$, $Y_{\text{n}}^{(t)}$.
\begin{equation}
    d^{(t)}_n = \| Y^{t}_{0} - Y^{t}_{n} \|_2
\end{equation}
Let $\tau$ denote a safe distance threshold and let $\epsilon$ be a buffer region. For each distance value $d^{(t)}_n$, we define the following penalty function:
\begin{equation}
    \phi(d)=
        \begin{cases}
        -(d - \tau) + \frac{\epsilon}{2} & (d < \tau) \\
        \dfrac{(d - \tau - \epsilon)^2}{2\epsilon} & (\tau \le d < \tau + \epsilon) \\
        0 & (d \ge \tau + \epsilon).
        \end{cases}
\end{equation}
A linear penalty is applied when the ego-neighbor distance falls below the threshold $\tau$, and a quadratic transition region is used to smoothly interpolate between the linear penalty and the zero-penalty region within the interval $\tau \le d < \tau + \epsilon$.
When the distance is sufficiently large, the penalty becomes zero.
The total collision cost for each neighbor agent $n$ is obtained by accumulating the penalty over time:
\begin{equation}
    C_n = \sum_{t=1}^{T_{\text{fut}}}\phi(d^{(t)}_n)
\end{equation}
To focus on the neighbor that poses the highest risk to the ego agent, the final collision cost is defined as
\begin{equation}
    \mathcal{C}(Y_{\text{plan}}, Y_{\text{pred}})=\max_{n}C_n
\end{equation}
This worst-case collision cost is averaged across the batch and used as part of the final loss.

\section{Evaluation Metrics}
\label{sup:evaluation_metrics}
\begin{enumerate}
\item \textbf{Average Displacement Error (ADE)}
ADE evaluates how closely the model-generated future path of the agent matches the ground-truth trajectory in the dataset, based on the L2 distance. For the planning task, ADE is computed only for the ego agent, while for the prediction task, ADE is reported as the average over all non-ego agents in the scene. The formal definition of the ADE metric is provided below:
\begin{equation}
    \text{ADE} = \frac{1}{NT_{fut}} \sum_{i=1}^N \sum_{j=1}^{T_{fut}} \|\textbf{x}_i^{j} - \textbf{g}_i^{j}\|
\end{equation}
\noindent where $N$ is the number of samples, $\mathbf{x}$ is the generated path, and $\mathbf{g}$ is the ground truth path.

\item \textbf{Collision Rate (CR)}
Collision Rate is an essential metric for ensuring safe autonomous driving. A collision is considered to occur when any point in the model-generated plan of the ego agent comes within a specified threshold distance from the ground-truth future path of surrounding agents. Following the evaluation setup used in GameTheoretic~\cite{supp-kedia2023GameTheoretic}, we set the collision threshold to 0.6. The formal definition of the Collision Rate metric is provided below:
\begin{equation}
     \text{CR} = \frac{1}{N} \sum_{i=1}^{N} \mathbf{1}\left( \min_{j=1,\dots,T_{fut}} \|\mathbf{x}_i^{j} - \mathbf{g_{sur}}_i^{j}\| < {\epsilon}_{c} \right)
\end{equation}
\noindent where ${\epsilon}_{c}$ is the collision threshold, and $\mathbf{g_{sur}}$ is the surrounding agent's ground-truth future path.

    \item \textbf{Final Displacement Error (FDE)}
FDE evaluates how far the final point of the generated trajectory deviates from the final point of the ground-truth trajectory, computed based on the L2 distance. As with ADE, FDE is computed only for the ego agent in the planning task, while in the prediction task it is calculated for all surrounding agents and then averaged for reporting. The formal definition of the FDE metric is provided below:
\begin{equation}
    \text{FDE} = \frac{1}{N} \sum_{i=1}^N \|\textbf{x}_i^{T_{fut}} - \textbf{g}_i^{T_{fut}}\|
\end{equation}

\item \textbf{Miss Rate (MR)}
Miss Rate evaluates the similarity of the final point in a manner similar to FDE, but in a discrete form: if the final point of the generated plan is farther than a specified threshold from the ground-truth final point, it is counted as a miss. Following the experimental setup in~\cite{supp-note2eplan_3_ICCV2025_lee2025interaction}, we set the miss-rate threshold to 0.5. The formal definition of the Miss Rate metric is provided below:
\begin{equation}
    \text{MR} = \frac{1}{N} \sum_{i=1}^N \mathbf{1}(\|\textbf{x}_i^{T_{fut}} - \textbf{g}_i^{T_{fut}}\| > {\epsilon}_{m})
\end{equation}
\noindent where ${\epsilon}_{m}$ is the miss-rate threshold.

\end{enumerate}

\section{Implementation Details}
\label{sup:Implemenrtation_details}
\begin{enumerate}
\item \textbf{JRDB Dataset}
We utilized an NVIDIA RTX 3090 GPU for training models on the JRDB dataset and all training stages used the Adam optimizer ($\beta_1=0.9,\beta_2=0.99$). Prior to employing our proposed DPT method, we trained a unified pretrained model using a game-theoretic loss. The hyperparameters for pretraining were set as follows: a batch size of 16, 200 epochs, and a learning rate of $1 \times 10^{-4}$. For fine-tuning on individual tasks without DPT (initialized from the pretrained model), we used a batch size of 16, 100 epochs, and a learning rate of $1 \times 10^{-5}$. For the DPT model, we utilized a batch size of 64, with both the allocation and training phases set to 50 epochs each. Additionally, we set the planner and prediction allocation ratios to 50\%. The learning rates were set to $5 \times 10^{-6}$ for the planner and $2 \times 10^{-6}$ for the predictor.

\item \textbf{JTA Dataset}
Model training on the JTA dataset was performed using an NVIDIA V100 GPU. Similar to the JRDB setup, we trained a unified pretrained model using the game-theoretic loss before applying DPT and also adopted the Adam optimizer ($\beta_1=0.9, \beta_2=0.99$). The hyperparameters were set to a batch size of 16, 30 epochs, and a learning rate of $1 \times 10^{-4}$. Fine-tuning for individual tasks without DPT was conducted with a batch size of 4, 30 epochs, and a learning rate of $1 \times 10^{-5}$. The DPT model utilized a batch size of 8, with both allocation and training phases set to 15 epochs. The planner and prediction allocation ratios were both set to 50\%, and the learning rates for the planner and predictor were both set to $1 \times 10^{-5}$.

\end{enumerate}

\section{Qualitative Results}
\label{sup:Qualitative_results}
\begin{figure*}[t]
    \centering
    \includegraphics[width=\textwidth]{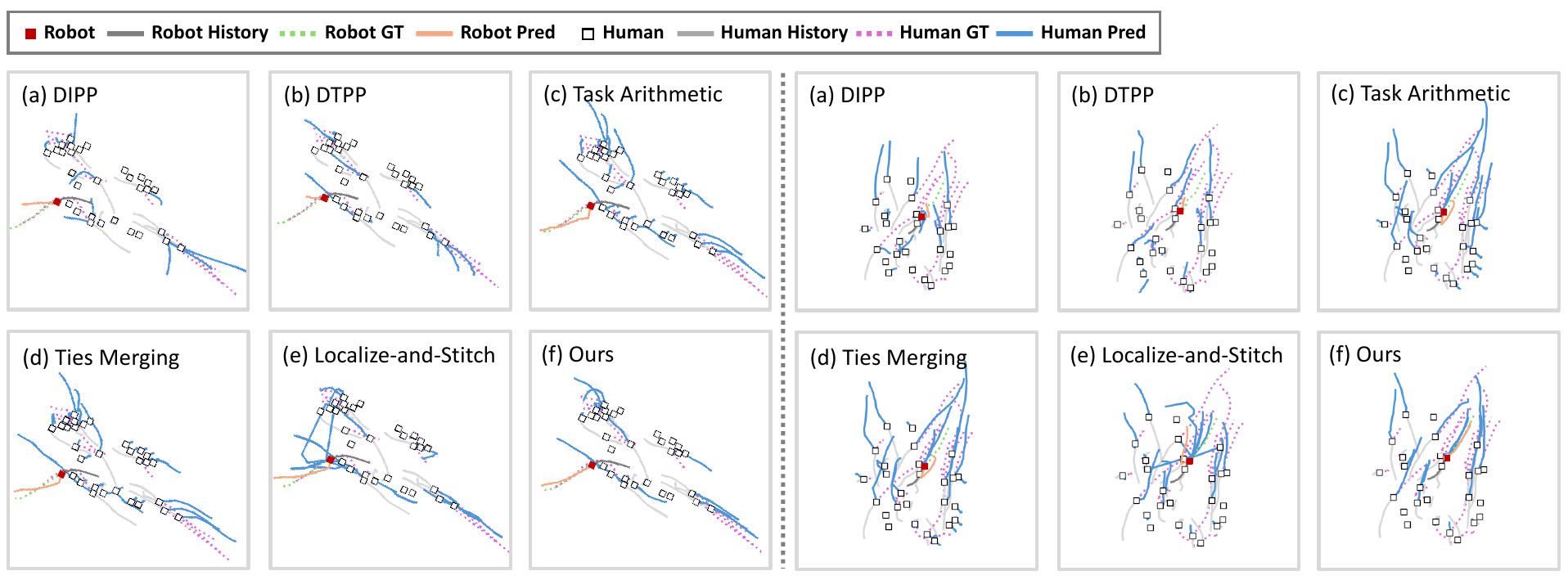}
    \caption{Qualitative Results on the JRDB dataset. (a)–(b) show unified models with different architectures, while (c)–(e) present results of applying various baseline model merging methods on our unified model. These are compared with (f) Ours, which employs DPT and Sparse Merging. In both examples, our method accurately predicts surrounding agents and produces safer plans compared to the baselines.}
    \label{fig:sup_Qualitative_results}
\end{figure*}
We provide additional qualitative results demonstrating the effectiveness of our DPT + Sparse Merging approach. We apply a DPT allocation ratio of 50:50 and use a \textit{mask ratio} of $K=1$. Fig.~\ref{fig:sup_Qualitative_results} presents results on the JRDB dataset, showing that our method performs robustly in dense scenes.

\begin{figure}[htbp]
    \centering
    \includegraphics[width=0.8\columnwidth]{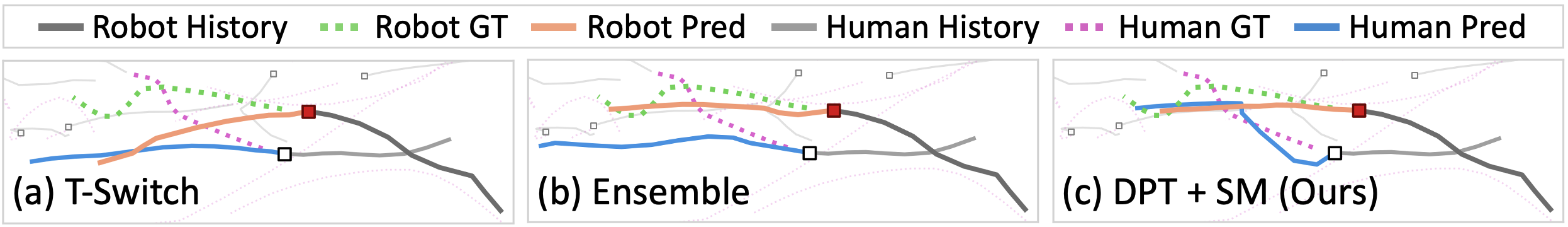}
    \caption{Complementary Qualitative Results on the JRDB.}
    \label{fig:qualitative_interaction}
\end{figure}
Fig.~\ref{fig:qualitative_interaction} complements other qualitative results with an additional interaction case. Planning's safety-oriented representation degrades prediction, so (a) T-Switch and (b) Ensemble predict trajectories that drift apart even in robot–human interaction scenarios. DPT + SM instead alleviates this task conflict and preserves the interaction pattern (c).

\section{Empirical Study}
\label{sup:Empirical_Study}

\subsection{Skill Conflict in Other Architecture}
\begin{figure}[t]
    \centering
    \includegraphics[width=0.95\textwidth]{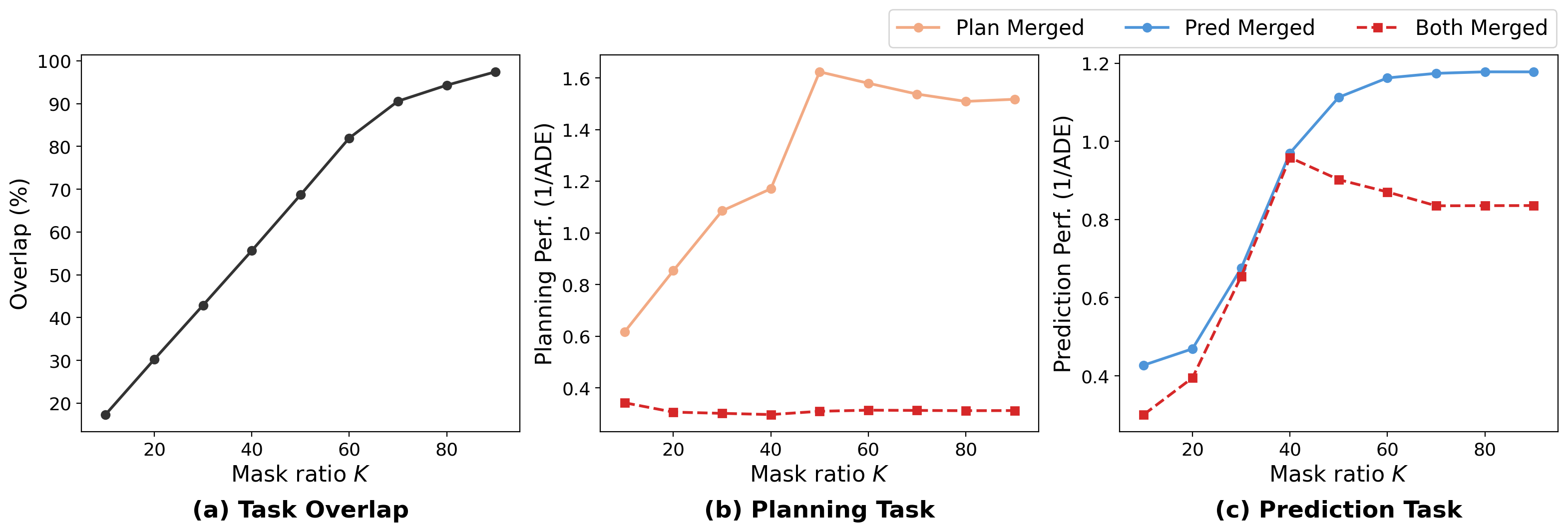}
    \caption{(a) Parameter overlap ratio between planning and prediction tasks. (b)-(c) Planning and prediction performance for single-task merging and joint-task merging on the DTPP architecture.}
    \label{fig:DTPP}
    \vspace{10pt}
\end{figure}

\begin{figure}[t]
    \centering
    \includegraphics[width=0.9\textwidth]{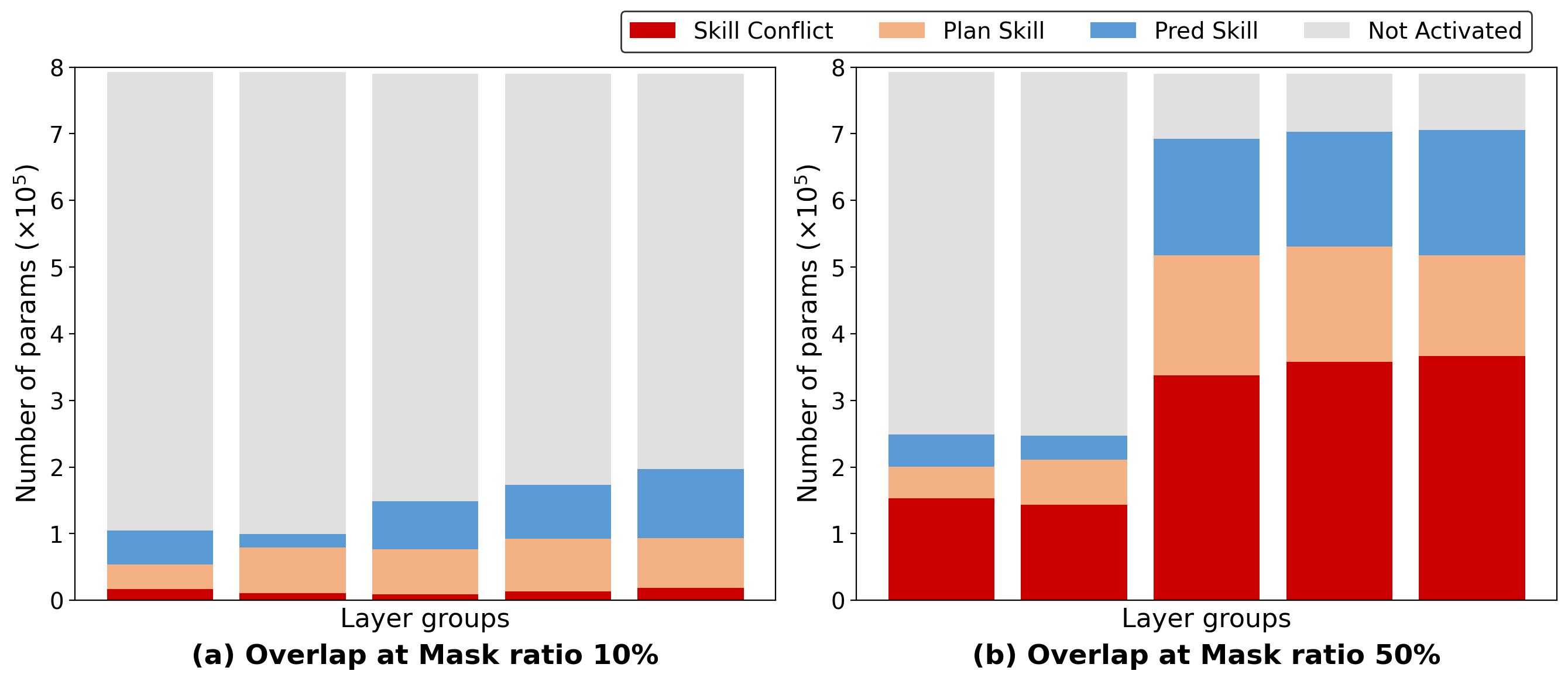}
    \caption{Parameter distribution by layer group and the degree of skill conflict between Plan Skill and Pred Skill in the DTPP architecture at mask ratios of $K{=}10\%$ and $K{=}50\%$.}
    \label{fig:DTPP_Overlap}
\end{figure}

\begin{figure}[htbp]
    \centering
    \includegraphics[width=0.5\textwidth]{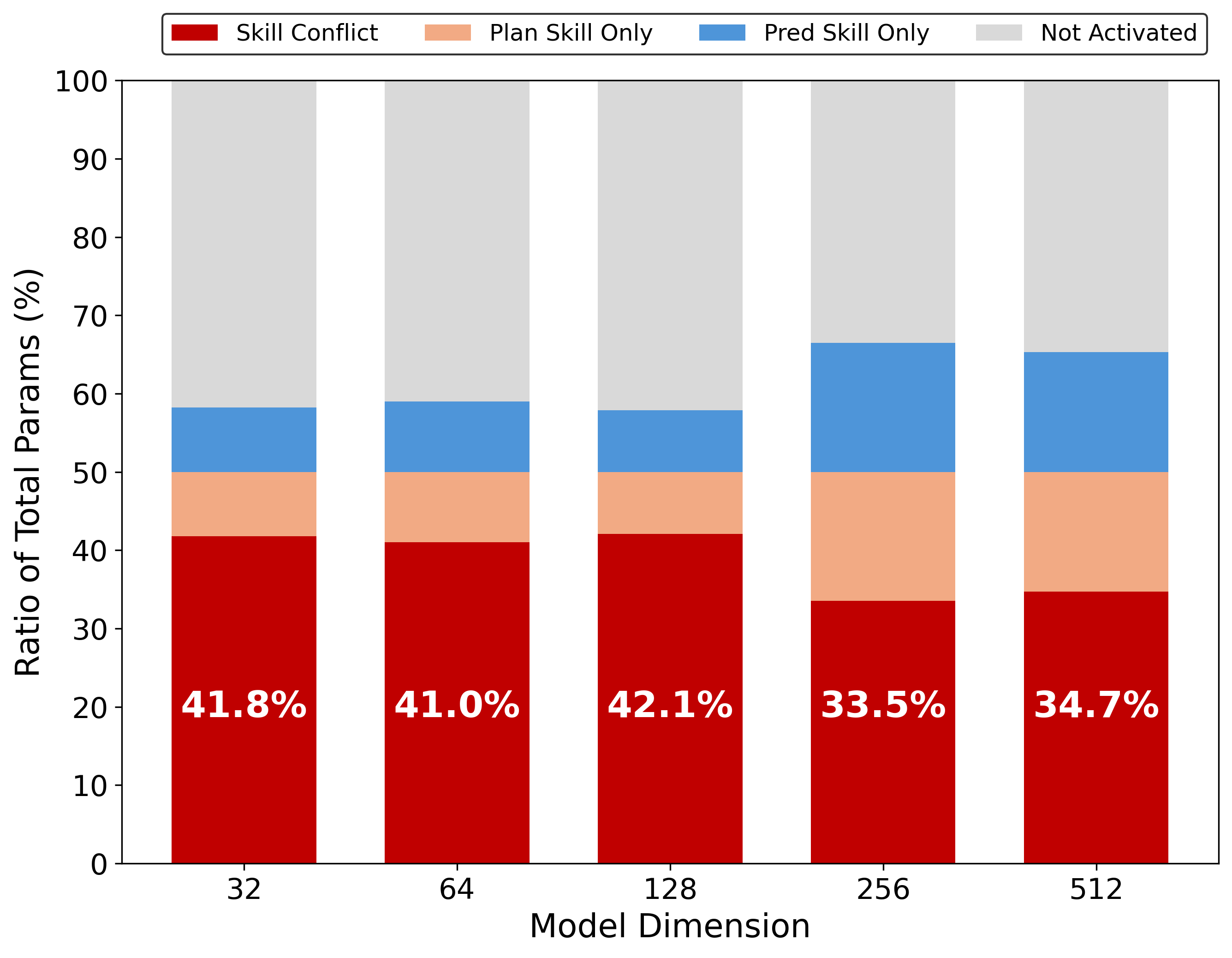}
    \caption{Effect of model dimension on skill conflict. The overlap ratio is measured under a fixed mask ratio of $K{=}50\%$. Compact models with dimensions from 32 to 128 exhibit consistently high overlap ratios of 41--42\%, while larger models with dimensions of 256 and 512 show reduced overlap ratios of 34--35\%.}
    \label{fig:sup_fig_model_dim}
\end{figure}

To verify that the Skill Conflict analysis conducted in Sec. 3.3 of the main paper is not a phenomenon limited to a specific model architecture, we present an identical empirical study using the DTPP architecture \cite{supp-huang2024DTPP}, which is another unified model backbone.
As shown in Fig.~\ref{fig:DTPP} (a), the parameter overlap ratio between the planning and prediction tasks increases as the mask ratio $K$ increases. In the performance graphs for the (b) planning task and (c) prediction task, it is evident that when merging only a single task (Plan Merged, Pred Merged), performance improves as the parameter utilization ratio increases. However, in the two-task merging scenario (Both Merged), performance degradation occurs due to the interference arising from overlapping activated parameter regions.
Fig.~\ref{fig:DTPP_Overlap} indicates the parameter distribution and the degree of Skill Conflict within layer groups. At a low mask ratio of 10\% (a), the parameter regions utilized by the two tasks are largely disjoint, resulting in minimal Skill Conflict, with $\text{Overlap}(10\%)=1.73\%$. When the mask ratio increases to 50\% (b), both tasks activate significantly wider areas within the same layer groups, leading to a substantial increase in Skill Conflict, with $\text{Overlap}(50\%)=34.34\%$.
Consequently, the empirical results in Fig.~\ref{fig:DTPP} and Fig.~\ref{fig:DTPP_Overlap} obtained from the DTPP backbone suggest that the Skill Conflict phenomenon is not an issue unique to the Social-Transmotion backbone \cite{supp-TP_39_ICLR2024_saadatnejad2023social}. Rather, it is a pervasive phenomenon observed across unified models with limited parameter capacity.

\subsection{Effect of Model Capacity on Skill Conflict}
\label{sec:Effect_of_Model_Capacity_on_Skill_Conflict}

To analyze how model capacity affects skill conflict between the prediction and planning tasks, we vary the backbone model dimension while measuring the conflict ratio under a fixed mask ratio of $K{=}50\%$.
The conflict ratio is defined as the proportion of parameters that are simultaneously activated by both tasks.
Specifically, for each task we compute the parameter update magnitude relative to the base model and define the top 50\% parameters as activated parameters.
As shown in Fig.~\ref{fig:sup_fig_model_dim}, compact models with dimensions ranging from 32 to 128 exhibit consistently high conflict ratios of approximately 41--42\%.
In contrast, larger models with dimensions of 256 and 512 show reduced conflict ratios of around 33--34\%.
These results suggest that when model capacity is limited, the two tasks tend to rely on highly overlapping parameter regions, leading to larger interference in the parameter space.
On the other hand, larger models provide more capacity to allocate task-specific parameter spaces, which alleviates the degree of conflict.
Overall, these findings indicate that skill conflict becomes more pronounced in compact unified models, highlighting the need for mechanisms that explicitly control parameter interactions during training.

\begin{table}[htbp]
\centering
\caption{Parameter overlap under model size scaling on JRDB.}
\vspace{-1pt}
\label{tab:r4_capacity}
\setlength{\tabcolsep}{5pt}
\renewcommand{\arraystretch}{0.9}
\scriptsize
\begin{tabular*}{0.66\columnwidth}{@{\extracolsep{\fill}}lccc}
\toprule
\multirow{2}{*}{Params} & \multicolumn{3}{c}{\textit{Overlap}($K$) (\%)} \\
\cmidrule(lr){2-4}
& $K{=}1\%$ & $K{=}25\%$ & $K{=}50\%$ \\
\midrule
0.87\,M            & 1.75 & 20.40 & 42.58 \\
3.22\,M (default)  & 1.06 & 17.47 & 38.82 \\
7.04\,M            & 1.01 & 17.06 & 36.94 \\
\bottomrule
\end{tabular*}

\vspace{-12pt}
\end{table}
Capacity scaling alone does not auto-resolve the conflict in our regime. Following Sec.~\textcolor{eccvblue}{3.3} (Eq.~\textcolor{eccvblue}{9}), Tab.~\ref{tab:r4_capacity} shows that overlap decreases only mildly with model size but remains high for all $K$. Our data scale does not permit foundation-scale training, and within the trainable range increasing capacity fails to meaningfully reduce overlap, supporting the necessity of DPT's explicit overlap intervention.

\subsection{Sensitivity Analysis of Skill Conflict}
\label{sec:Sensitivity_Analysis_of_Skill_Conflict}

\begin{table}[htbp]
\centering
\caption{
Sensitivity of DPT scheduling on JRDB with Sparse Merging ($K{=}1\%$).
}
\vspace{-1pt}
\label{tab:dpt_phase_ratio}
\setlength{\tabcolsep}{4.2pt}
\renewcommand{\arraystretch}{1.0}
\scriptsize

\resizebox{0.9\columnwidth}{!}{%
\begin{tabular}{c|cc|cccc|cc}
\toprule
\multicolumn{3}{c|}{DPT Scheduling}
& \multicolumn{4}{c|}{Planning Metric}
& \multicolumn{2}{c}{Prediction Metric} \\
\cmidrule(lr){1-3}\cmidrule(lr){4-7}\cmidrule(lr){8-9}
Ratio & Mask Growth & Mask Frozen
& ADE $\downarrow$ & Col. $\downarrow$ & FDE $\downarrow$ & MR $\downarrow$
& ADE $\downarrow$ & FDE $\downarrow$ \\
\midrule
1:3 & $1$ to $E/4$ & $E/4$ to $E$
& 0.4328 & 0.0098 & 0.7704 & 0.3779 & 0.5990 & 1.0402 \\

1:1 & $1$ to $E/2$ & $E/2$ to $E$
& 0.4044 & 0.0091 & 0.7458 & 0.3706 & 0.5952 & 1.0352 \\

3:1 & $1$ to $3E/4$ & $3E/4$ to $E$
& 0.4110 & 0.0099 & 0.7368 & 0.3706 & 0.5740 & 1.0108 \\
\bottomrule
\end{tabular}%
}
\vspace{-12pt}
\end{table}

\begin{table}[htbp]
\centering
\caption{
DPT allocation ratio sweep on full JTA.
}
\vspace{-1pt}
\label{tab:jta_dpt_allocation}

\setlength{\tabcolsep}{4.2pt}
\renewcommand{\arraystretch}{0.9}
\scriptsize
\resizebox{0.9\columnwidth}{!}{
\begin{tabular}{cc|cccc|cc}
\toprule
\multicolumn{2}{c|}{DPT Allocation Ratio}
& \multicolumn{4}{c|}{Planning Metric}
& \multicolumn{2}{c}{Prediction Metric} \\
\cmidrule(lr){1-2}\cmidrule(lr){3-6}\cmidrule(lr){7-8}
Plan (\%) & Pred (\%)
& ADE $\downarrow$ & Col. $\downarrow$ & FDE $\downarrow$ & MR $\downarrow$
& ADE $\downarrow$ & FDE $\downarrow$ \\
\midrule

90 & 10
& \cellcolor{blue!10}{0.9631}
& \cellcolor{blue!10}{0.0054}
& \cellcolor{blue!10}{1.8385}
& \cellcolor{blue!20}{0.8074}
& \cellcolor{red!45}{\textbf{1.0765}}
& \cellcolor{red!30}{2.0967} \\

75 & 25
& \cellcolor{blue!20}{0.9589}
& \cellcolor{blue!37}{0.0053}
& \cellcolor{blue!20}{1.8363}
& \cellcolor{blue!10}{0.8170}
& \cellcolor{red!40}{1.0824}
& \cellcolor{red!20}{2.1009} \\

50 & 50
& \cellcolor{blue!30}{0.9435}
& \cellcolor{blue!45}{\textbf{0.0052}}
& \cellcolor{blue!30}{1.8335}
& \cellcolor{blue!30}{0.8042}
& \cellcolor{red!35}{1.0863}
& \cellcolor{red!45}{\textbf{2.0930}} \\

25 & 75
& \cellcolor{blue!42}{0.9125}
& \cellcolor{blue!37}{0.0053}
& \cellcolor{blue!42}{1.8114}
& \cellcolor{blue!45}{\textbf{0.8010}}
& \cellcolor{red!20}{1.1153}
& \cellcolor{red!15}{2.1354} \\

10 & 90
& \cellcolor{blue!45}{\textbf{0.9059}}
& \cellcolor{blue!37}{0.0053}
& \cellcolor{blue!45}{\textbf{1.8020}}
& \cellcolor{blue!42}{0.8012}
& \cellcolor{red!10}{1.1252}
& \cellcolor{red!10}{2.1553} \\

\bottomrule
\end{tabular}%
}
\end{table}

\begin{table}[htbp]
\centering
\caption{
Robustness of our method across three seeds on JRDB.
}
\vspace{-1pt}
\label{tab:dpt_seed_robustness}
\setlength{\tabcolsep}{4.2pt}
\renewcommand{\arraystretch}{1.0}
\scriptsize
\resizebox{\columnwidth}{!}{%
\begin{tabular}{l|cccc|cc}
\toprule
\multirow{2}{*}[-0.6ex]{Method}
& \multicolumn{4}{c|}{Planning Metric}
& \multicolumn{2}{c}{Prediction Metric} \\
\cmidrule(lr){2-5}
\cmidrule(lr){6-7}
& ADE $\downarrow$ & Col. $\downarrow$ & FDE $\downarrow$ & MR $\downarrow$ & ADE $\downarrow$ & FDE $\downarrow$ \\
\midrule
DPT + SM
& $0.4086{\pm}0.0081$ & $0.0094{\pm}0.0005$ & $0.7452{\pm}0.0021$ & $0.3721{\pm}0.0013$ & $0.5898{\pm}0.0046$ & $1.0305{\pm}0.0040$ \\
DPT + SM + JR
& $0.3655{\pm}0.0013$ & $0.0073{\pm}0.0001$ & $0.7145{\pm}0.0035$ & $0.3620{\pm}0.0027$ & $0.5168{\pm}0.0061$ & $0.9526{\pm}0.0113$ \\
\bottomrule
\end{tabular}%
}
\end{table}

To verify that the proposed DPT method operates robustly with respect to mask allocation scheduling, task allocation ratios, and random initialization, we conduct a series of sensitivity analyses.

Tab.~\ref{tab:dpt_phase_ratio} analyzes the sensitivity of DPT to different mask growth and mask freezing schedules. Specifically, we vary the ratio between the mask allocation phase and the mask freezing phase as 1:3, 1:1, and 3:1. As shown in Tab.~\ref{tab:dpt_phase_ratio}, the performance differences across different schedules are relatively small, indicating that DPT does not overly depend on a particular scheduling strategy.

Tab.~\ref{tab:jta_dpt_allocation} investigates the effect of varying the parameter allocation ratio between planning and prediction on the full JTA dataset. We observe a clear trade-off that allocating more parameters to planning improves planning performance while degrading prediction performance. This trend is consistent with the observations on JRDB, suggesting that the optimal allocation ratio is primarily determined by task priorities rather than dataset characteristics.

Tab.~\ref{tab:dpt_seed_robustness} reports the performance of our method across three different random seeds. The standard deviations are consistently small across all evaluation metrics, demonstrating that both DPT and DPT+SM+JR are robust to random seeds and provide stable performance.

\subsection{Causal Analysis of Skill Conflict}
\label{sec:Causial_Analysis_of_Skill_Conflict}

\begin{table}[htbp]
\centering
\caption{
Causal link between overlap and performance on JRDB.
}
\label{tab:forced_overlap}

\setlength{\tabcolsep}{4.2pt}
\renewcommand{\arraystretch}{0.9}
\scriptsize
\resizebox{0.8\columnwidth}{!}{
\centering
\begin{tabular}{c|cccc|cc}
\toprule
\multirow{2}{*}[-0.6ex]{Overlap (\%)}
& \multicolumn{4}{c|}{Planning Metric}
& \multicolumn{2}{c}{Prediction Metric} \\
\cmidrule(lr){2-5}\cmidrule(lr){6-7}
& ADE $\downarrow$ & Col. $\downarrow$ & FDE $\downarrow$ & MR $\downarrow$ & ADE $\downarrow$ & FDE $\downarrow$ \\
\midrule
0
& \cellcolor[HTML]{E89A8A}$\mathbf{0.4230}$
& \cellcolor[HTML]{E89A8A}$\mathbf{0.0101}$
& \cellcolor[HTML]{E89A8A}$\mathbf{0.7410}$
& \cellcolor[HTML]{E89A8A}$\mathbf{0.3745}$
& \cellcolor[HTML]{E89A8A}$\mathbf{0.5895}$
& \cellcolor[HTML]{E89A8A}$\mathbf{1.0505}$ \\
5
& \cellcolor[HTML]{F2B5A8}$0.4645$
& \cellcolor[HTML]{F2B5A8}$0.0110$
& \cellcolor[HTML]{F2B5A8}$0.7937$
& \cellcolor[HTML]{FCF1ED}$0.3839$
& \cellcolor[HTML]{F2B5A8}$0.5973$
& \cellcolor[HTML]{F9D5CE}$1.0620$ \\
10
& \cellcolor[HTML]{F9D5CE}$0.4704$
& \cellcolor[HTML]{F9D5CE}$0.0113$
& \cellcolor[HTML]{F9D5CE}$0.7976$
& \cellcolor[HTML]{FDEEEC}$0.3823$
& \cellcolor[HTML]{F9D5CE}$0.5989$
& \cellcolor[HTML]{F2B5A8}$1.0592$ \\
20
& \cellcolor[HTML]{FDEEEC}$0.4758$
& \cellcolor[HTML]{FDEEEC}$0.0114$
& \cellcolor[HTML]{FDEEEC}$0.7983$
& \cellcolor[HTML]{F2B5A8}$0.3816$
& \cellcolor[HTML]{FDEEEC}$0.6334$
& \cellcolor[HTML]{FDEEEC}$1.0877$ \\
30
& \cellcolor[HTML]{FCF1ED}$0.4819$
& \cellcolor[HTML]{FCF1ED}$0.0116$
& \cellcolor[HTML]{FCF1ED}$0.8013$
& \cellcolor[HTML]{F2B5A8}$0.3816$
& \cellcolor[HTML]{FCF1ED}$0.6487$
& \cellcolor[HTML]{FCF1ED}$1.0987$ \\
\bottomrule
\end{tabular}
}
\vspace{-12pt}
\end{table}

Our central hypothesis is that parameter overlap between planning and prediction induces Skill Conflict, which leads to performance degradation. To validate this hypothesis, we conduct an experiment in which a controlled amount of overlap is forcibly introduced between the masks of the two tasks while keeping all other training settings fixed.
Tab.~\ref{tab:forced_overlap} reports the performance as the overlap ratio is gradually increased from 0\% to 30\%. As the overlap ratio increases, we observe a consistent degradation in most planning and prediction metrics. In particular, both planning ADE and prediction ADE exhibit a clear performance decline as the amount of overlap increases.
These results go beyond a simple correlation and suggest that parameter overlap itself acts as a direct causal factor of performance degradation. Therefore, mitigating Skill Conflict requires mechanisms that explicitly control parameter overlap across tasks, such as DPT.

\subsection{Skill Conflict Beyond Planning and Prediction}
\label{sec:Skill_Conflict_Beyond_Planning_and_Prediction}

\begin{table}[htbp]
\centering
\caption{Sub-tasks overlap and 2-way vs. 3-way DPT on JRDB.}
\vspace{-1pt}
\label{tab:moresplit}

\setlength{\tabcolsep}{4.2pt}
\renewcommand{\arraystretch}{0.95}
\scriptsize
\resizebox{1.0\columnwidth}{!}{%
\begin{tabular}{l c c c c c c c}
\toprule
 & Pred overlap & \multicolumn{4}{c}{Planning Metric} & \multicolumn{2}{c}{Prediction Metric} \\
\cmidrule(lr){3-6} \cmidrule(lr){7-8}
Task split & $K{=}1\%$ & ADE$\downarrow$ & Col.$\downarrow$ & FDE$\downarrow$ & MR$\downarrow$ & ADE$\downarrow$ & FDE$\downarrow$ \\
\midrule

plan / pred-fast / pred-slow \textit{\scriptsize(velocity)}
& 6.90\% & 0.4947 & 0.0149 & 0.8002 & 0.3859 & 0.6872 & 1.1428 \\

plan / pred-near / pred-far \textit{\scriptsize(distance from ego)}
& 12.99\% & 0.4240 & 0.0107 & 0.7656 & 0.3790 & 0.6039 & 1.0508 \\

plan / pred-dense / pred-sparse \textit{\scriptsize(crowd density)}
& \textbf{15.13\%} & 0.4739 & 0.0151 & 0.7788 & 0.3818 & \textbf{0.5947} & 1.0353 \\

\midrule

plan / pred \textit{\scriptsize(paper, 2-way)}
& --- & \textbf{0.4044} & \textbf{0.0091} & \textbf{0.7458} & \textbf{0.3706} & 0.5952 & \textbf{1.0352} \\

\bottomrule
\end{tabular}
}
\end{table}

To investigate whether Skill Conflict is a more general phenomenon in multi-task learning, we further decompose the prediction task into finer-grained sub-tasks. Specifically, the original prediction task is split into two sub-tasks based on three criteria: (1) velocity, (2) distance from the ego agent, and (3) crowd density. Together with planning, this results in a total of three tasks, on which we perform 3-way DPT.
As shown in Tab.~\ref{tab:moresplit}, a substantial amount of parameter overlap is still observed among the prediction sub-tasks. This finding suggests that Skill Conflict is not limited to the interaction between planning and prediction, but can also arise within the prediction task itself.
Interestingly, larger overlaps among prediction sub-tasks lead to relatively better prediction performance. However, as the number of tasks increases, more tasks are forced to share a limited parameter region, thereby increasing the capacity cost. Consequently, our results indicate that the 2-way decomposition, plan–pred 2-way split, is the sweet spot.

\subsection{Analysis of Evolution of Disjoint Masks}
\label{sec:Anaylsis_of_Evolution_of_Disjoint_Masks}

\begin{figure}[htbp]
    \centering
    \includegraphics[width=0.8\columnwidth]{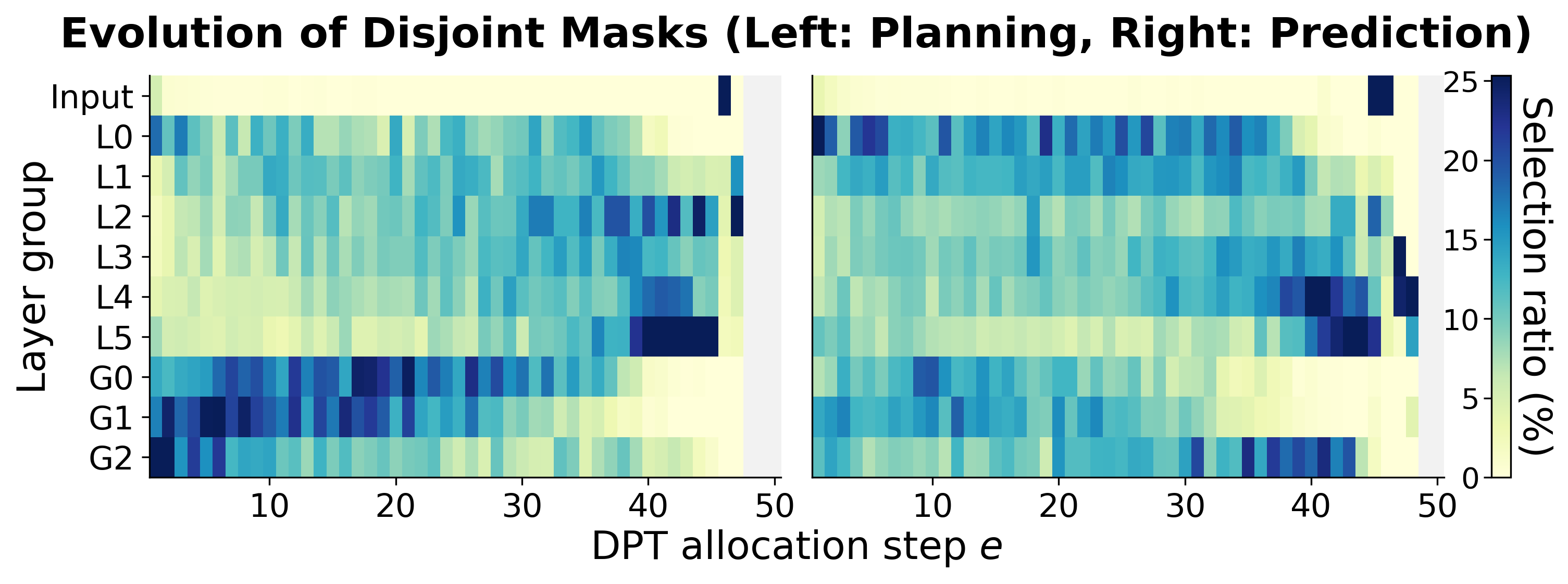}
    \caption{Evolution of disjoint masks across DPT allocation steps.}
    \label{fig:layer_allocation}
\end{figure}

Figure~\ref{fig:layer_allocation} shows how the disjoint masks evolve during DPT steps. Planning is initially concentrated on the global layers (G0-G2) and gradually extends to the early layers (L0-L2), while prediction starts from a relatively uniform distribution and converges onto the middle layers (L3-L5) in later steps, yielding distinct dominant regions for each task.
Moreover, mask regions shift smoothly across steps, indicating stable alternating Top-K selection.

\subsection{Analysis of Skill Conflict per Layer}
\label{sec:Anaylsis_of_Skill_Conflict_per_Layer}

\begin{figure}[htbp]
    \centering
    \includegraphics[width=0.8\columnwidth]{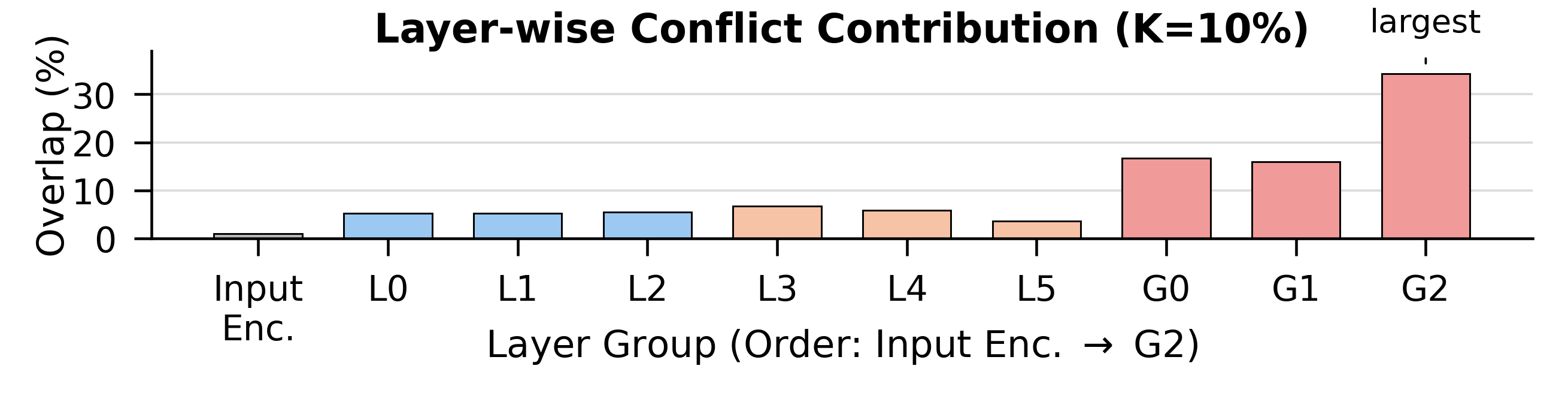}
    \caption{Layer-wise distribution of Skill Conflict.}
    \label{fig:layer_conflict}
\end{figure}

Fig.~\ref{fig:layer_conflict} indicates the layer-wise distribution of Skill Conflict. We observe that the overlap ratio consistently increases as the network depth becomes larger. In particular, the highest degree of overlap occurs in the deep global interaction layers, which are responsible for modeling high-level interaction information. These findings suggest that task conflicts in unified architectures tend to concentrate in specific layers, especially in deeper layers.

\section{Additional Baseline Comparisons}
\label{sup:Additional_Baseline_Comparisons}

\begin{table}[htbp]
\centering
\caption{Applicability of DPT to T-Switch on the JRDB dataset. The upper rows show task-specific fine-tuned models, and the lower rows show the effect of applying T-Switch with and without DPT.
}
\label{tab:tswitch_dpt}
\setlength\tabcolsep{6.5pt} %
\renewcommand{\arraystretch}{1.15}
\vspace{-4pt}
\resizebox{\textwidth}{!}{
\begin{tabular}{l|cccc|cc}
\hline
\multirow{2}{*}{Method}
& \multicolumn{4}{c|}{Planning Metric}
& \multicolumn{2}{c}{Prediction Metric} \\
\cline{2-7}
& \makecell{Effectiveness \\ (ADE $\downarrow$)}
& \makecell{Safety \\ (Col. rate $\downarrow$)}
& \makecell{Goal Success \\ (FDE $\downarrow$)}
& \makecell{Goal Success \\ (Miss rate $\downarrow$)}
& \makecell{ADE $\downarrow$}
& \makecell{FDE $\downarrow$} \\
\hline

Plan Finetune
& 0.3965 & 0.0078 & 0.7371 & 0.3700 & 1.9825 & 2.4087 \\
\rowcolor[HTML]{EFEFEF}
Plan Finetune + DPT
& 0.3699 & 0.0073 & 0.7244 & 0.3600 & 0.7368 & 1.1702 \\

Pred Finetune
& 0.8889 & 0.0195 & 1.2140 & 0.9615 & 0.4967 & 0.9144 \\
\rowcolor[HTML]{EFEFEF}
Pred Finetune + DPT
& 0.8469 & 0.0200 & 1.1244 & 0.9189 & 0.5250 & 0.9682 \\
\midrule

T-Switch\cite{supp-Merging_TSwitch_qi2025less}
& 0.8047 & 0.0199 & 1.1009 & 0.8784 & 0.8091 & 1.2112 \\
\rowcolor[HTML]{DAE8FC}
T-Switch + DPT
& \textbf{0.5894} & \textbf{0.0148} & \textbf{0.8975} & \textbf{0.4455} & \textbf{0.6230} & \textbf{1.0523} \\
\hline
\end{tabular}
}
\end{table}

As shown in Tab. \ref{tab:tswitch_dpt}, combining DPT with T-Switch improves planning and prediction metrics compared to using T-Switch alone on the JRDB dataset. These results demonstrate that DPT can improve the performance and stability of unified models when combined with various merging methods. They further indicate that binary switching-based merging approaches, such as T-Switch, are insufficient to fully resolve the skill conflict arising in the shared encoder, highlighting the importance of training-based parameter region separation such as DPT.

\begin{table}[htbp]
\centering
\caption{
Comparison with MTL baselines on JRDB.
}
\label{tab:MTL_Baseline}
\setlength\tabcolsep{6.5pt}
\renewcommand{\arraystretch}{1.0}
\vspace{-4pt}
\resizebox{0.7\textwidth}{!}{
\begin{tabular}{l|cccc|cc}
\toprule
Method
& \multicolumn{4}{c|}{Planning Metric}
& \multicolumn{2}{c}{Prediction Metric} \\
\cmidrule(lr){2-5}\cmidrule(lr){6-7}
& ADE $\downarrow$ & Col. $\downarrow$ & FDE $\downarrow$ & MR $\downarrow$ & ADE $\downarrow$ & FDE $\downarrow$ \\
\midrule
PCGrad
& 0.3998 & \underline{0.0068} & 0.7862 & 0.3720 & 0.6726 & 1.1216 \\
CAGrad
& \underline{0.3861} & \textbf{0.0067} & 0.7629 & \underline{0.3657} & 0.6801 & 1.1387 \\
\rowcolor[HTML]{DAE8FC}
DPT + SM
& 0.4044 & 0.0091 & \underline{0.7458} & 0.3706 & \underline{0.5952} & \underline{1.0352} \\
\rowcolor[HTML]{DAE8FC}
DPT + SM + JR
& \textbf{0.3640} & 0.0074 & \textbf{0.7105} & \textbf{0.3589} & \textbf{0.5097} & \textbf{0.9396} \\
\bottomrule
\end{tabular}%
}
\end{table}

Representative multi-task learning (MTL) methods, such as PCGrad\cite{supp-yu2020gradient} and CAGrad\cite{supp-liu2021conflict}, have been proposed to mitigate gradient conflicts among tasks. PCGrad resolves conflicting gradient directions by projecting task gradients onto each other, whereas CAGrad seeks a balanced optimization direction by aggregating gradients from multiple tasks.
\textit{Gradient Conflict} and \textit{Skill Conflict} are similar in that both concern task interference within a single backbone. However, \textit{Gradient Conflict} focuses on the clash of gradient directions between tasks during joint training of a single model, whereas \textit{Skill Conflict} refers to the parameter-level overlap of each task's core regions in the shared encoder, which emerges when independently fine-tuned material models are merged. This phenomenon was not pronounced in large foundation models with ample capacity, and our work is the first to diagnose and resolve it in the compact unified prediction–planning regime. In Tab.~\ref{tab:MTL_Baseline}, PCGrad and CAGrad improve planning but degrade prediction, with capacity skewed toward one task, supporting the necessity of parameter-level separation.

\begin{table}[htbp]
\centering
\caption{
Full JTA results with DPT and merging methods.
}
\label{tab:jta_full_results}

\setlength{\tabcolsep}{4.2pt}
\renewcommand{\arraystretch}{1.0}
\scriptsize
\resizebox{0.85\columnwidth}{!}{%
\begin{tabular}{l|cccc|cc}
\toprule
\multirow{2}{*}[-0.6ex]{Method}
& \multicolumn{4}{c|}{Planning Metric}
& \multicolumn{2}{c}{Prediction Metric} \\
\cmidrule(lr){2-5}\cmidrule(lr){6-7}
& ADE $\downarrow$ & Col. $\downarrow$ & FDE $\downarrow$ & MR $\downarrow$ & ADE $\downarrow$ & FDE $\downarrow$ \\
\midrule
Task Arithmetic
& 0.9780 & \textbf{0.0057} & 1.8141 & 0.8317 & 1.1690 & 2.1406 \\
\rowcolor[HTML]{DAE8FC}
Task Arithmetic + DPT
& \textbf{0.9724} & 0.0060 & \textbf{1.7820} & \textbf{0.7911} & \textbf{1.1332} & \textbf{2.1094} \\
Ties Merging
& 0.9897 & 0.0059 & \textbf{1.7927} & 0.8116 & 1.1957 & 2.1619 \\
\rowcolor[HTML]{DAE8FC}
Ties Merging + DPT
& \textbf{0.9658} & \textbf{0.0057} & 1.7954 & \textbf{0.7941} & \textbf{1.1172} & \textbf{2.0950} \\
Sparse Merging
& 0.9932 & 0.0058 & 1.8083 & 0.8030 & 1.2034 & 2.1970 \\
\rowcolor[HTML]{DAE8FC}
Sparse Merging + DPT
& \textbf{0.9203} & \textbf{0.0052} & \textbf{1.8062} & \textbf{0.7897} & \textbf{1.0727} & \textbf{2.0759} \\
\bottomrule
\end{tabular}%
}
\end{table}

Tab.~\ref{tab:jta_full_results} analyzes the performance changes obtained by applying DPT to various model merging methods, including Task Arithmetic, Ties Merging, and Sparse Merging, on the full JTA dataset. As shown in Tab.~\ref{tab:jta_full_results}, incorporating DPT overall improves both planning and prediction performance across the merging methods.

\section{Training Cost and Efficiency}
\label{sup:Training_Cost_Comparison}

\begin{table}[htbp]
\centering
\caption{Quantitative comparison of training cost and task performance across different baselines. Training cost is measured in GPU-hours.}
\label{tab:TTC_table}
\setlength\tabcolsep{6.5pt} %
\renewcommand{\arraystretch}{1.15}
\vspace{-4pt}
\resizebox{\textwidth}{!}{
\begin{tabular}{l|cccc|cc|c}
\hline
\multirow{2}{*}{Method}
& \multicolumn{4}{c|}{Planning Metric}
& \multicolumn{2}{c|}{Prediction Metric}
& \multirow{3}{*}{GPU-Hour} \\
\cline{2-7}
& \makecell{Effectiveness \\ (ADE $\downarrow$)}
& \makecell{Safety \\ (Col. rate $\downarrow$)}
& \makecell{Goal Success \\ (FDE $\downarrow$)}
& \makecell{Goal Success \\ (Miss rate $\downarrow$)}
& \makecell{ADE $\downarrow$}
& \makecell{FDE $\downarrow$} \\
\hline

Joint Training (Vanilla)
& 0.9269 & 0.0189 & 1.1701 & 0.9587 & 0.6552 & 1.0670 & 1.58 \\
Joint Training (Game)
& 0.8960 & 0.0188 & 1.1365 & 0.9503 & 0.6471 & 1.0508 & 2.90 \\
Ties Merging (Baseline)
& 0.8759 & 0.0167 & 1.1503 & 0.9019 & 0.9463 & 1.3412 & 3.38 \\
\rowcolor[HTML]{DAE8FC}
Sparse Merging + DPT
& \textbf{0.4044} & \textbf{0.0091} & \textbf{0.7458} & \textbf{0.3706} & \textbf{0.5952} & \textbf{1.0352} & 4.10\\

\hline
\end{tabular}
}
\end{table}

We compare the training cost of DPT against several baselines in Tab.~\ref{tab:TTC_table}, including vanilla joint training, game-theoretic joint training~\cite{supp-kedia2023GameTheoretic}, and TIES merging~\cite{supp-Merging_Tiesmerging_yadav2023ties}. GPU-hours are calculated by multiplying the total training time by the number of GPUs used. Although our method incurs approximately 1.2$\times$ higher GPU-hours than TIES merging due to the additional mask management required by DPT, it more effectively mitigates task conflict and achieves better performance. Moreover, because our method is integrated into a single model after training, it does not introduce additional inference-time cost compared with the other baselines. Thus, our method incurs additional cost only during training, while providing better performance without increasing the inference-time cost after deployment.

\begin{table}[htbp]
\centering
\caption{
Inference efficiency measured on an RTX 3090 with batch size 1.
}
\vspace{-1pt}
\label{tab:efficiency}
\setlength{\tabcolsep}{5pt}
\renewcommand{\arraystretch}{1.0}
\scriptsize
\resizebox{0.8\columnwidth}{!}{%
\begin{tabular}{ccccc}
\toprule
Latency (ms)
& GFLOPs $\downarrow$
& Params (M) $\downarrow$
& Model Size (MB) $\downarrow$
& Inference Memory (MB) $\downarrow$ \\
\midrule
10.4 & 1.65 & 6.44 & 24.55 & 1.96 \\
\bottomrule
\end{tabular}%
}
\end{table}

Tab.~\ref{tab:efficiency} reports the inference efficiency of the proposed model. All evaluations are conducted on an NVIDIA RTX 3090 GPU with a batch size of 1. Our model achieves a latency of 10.4 ms, requiring only 1.65 GFLOPs and 6.44M parameters. Moreover, it requires only 24.55 MB of model size and 1.96 MB of inference memory. This suggests that the model is deployable on edge devices such as NVIDIA Jetson Orin Nano (4GB/8GB).

\section{Adaptability on End-to-End Autonomous Driving}
\label{sup:Adaptability_on_E2EAD}

To validate the scalability of DPT and Sparse Merging, we extend our method, previously verified on prediction and planning tasks in the robotics domain, to end-to-end autonomous driving (E2E-AD) in the vehicle domain.
Early E2E-AD paradigms typically adopted a sequential architecture, where the output of each module is passed to the next module as its input~\cite{supp-E2E_39_old_jiang2023vad_endtoend3,supp-E2E_28_ICRA2025_sun2025sparsedrive,supp-E2E_42_old_hu2023planning_endtoend7,supp-liao2025diffusiondrive}. However, in such a structure, the planning loss has limited direct influence on tasks assigned to early layers due to the depth of the model, which can lead to meaningful correlations among tasks being overlooked. As a result, recent paradigms have introduced approaches that perform multiple tasks in parallel within a shared decoder rather than relying on a sequential structure~\cite{supp-E2E_41_old_weng2024drive_endtoend5,supp-E2E_25_ICLR2025_jia2025drivetransformer,supp-E2E_4_ICCV2025_tang2025hip}, achieving strong performance. Although this parallel design has been beneficial, existing models do not explicitly address the potential gradient conflicts among tasks in the shared decoder, which may be one of the reasons for their suboptimal performance.

\begin{table}[htbp]
\centering
\caption{Quantitative comparison on Bench2Drive across end-to-end autonomous driving models in closed-loop and open-loop evaluation, including HiP-AD and HiP-AD with DPT and Sparse Merging (SM).}
\label{tab:E2E_metric}

\setlength\tabcolsep{5.0pt}
\renewcommand{\arraystretch}{1.15}
\vspace{-4pt}

\resizebox{0.98\linewidth}{!}{%
\begin{tabular}{l|cc|cccc|cccc|ccc}
\toprule
\multirow{4}{*}[-0.8em]{Method}
& \multicolumn{2}{c|}{Closed-loop Metric}
& \multicolumn{11}{c}{Open-loop Metric} \\
\cmidrule(lr){2-3}\cmidrule(l){4-14}
& \multicolumn{2}{c|}{Planning Performance}
& \multicolumn{8}{c|}{Planning Performance}
& \multicolumn{3}{c}{Perception Performance} \\
\cmidrule(lr){2-3}\cmidrule(lr){4-11}\cmidrule(l){12-14}
& \multirow{2}{*}[-0.2em]{\makebox[1.8cm][c]{RC $\uparrow$}}
& \multirow{2}{*}[-0.2em]{\makebox[1.8cm][c]{DS $\uparrow$}}
& \multicolumn{4}{c|}{L2(m) $\downarrow$}
& \multicolumn{4}{c|}{Col. Rate(\%) $\downarrow$}
& \multirow{2}{*}[-0.2em]{mAP $\uparrow$}
& \multirow{2}{*}[-0.2em]{mATE $\downarrow$}
& \multirow{2}{*}[-0.2em]{NDS $\uparrow$} \\
\cmidrule(lr){4-7}\cmidrule(lr){8-11}
& & & 1s & 2s & 3s & Avg. & 1s & 2s & 3s & Avg. & & & \\
\midrule

VAD~\cite{supp-E2E_39_old_jiang2023vad_endtoend3}
& 68.24 & 52.36 & \textbf{0.45} & \underline{0.91} & \underline{1.48} & \textbf{0.95} & \underline{0.10} & \underline{0.20} & \underline{0.30} & \underline{0.20} & 0.5066 & 0.3852 & 0.6042 \\

UniAD~\cite{supp-E2E_42_old_hu2023planning_endtoend7}
& 66.34 & 53.42 & 0.52 & 1.27 & 1.69 & 1.16 & 0.77 & 3.85 & 7.93 & 4.18 & 0.1345 & 0.5204 & 0.3240 \\

DriveTransformer~\cite{supp-E2E_25_ICLR2025_jia2025drivetransformer}
& \underline{95.32} & 62.83 & 0.54 & 1.26 & 2.03 & 1.28 & 0.49 & 1.10 & 1.65 & 1.08 & 0.2835 & 0.5782 & 0.4888 \\

\rowcolor[gray]{0.93}
HiP-AD~\cite{supp-E2E_4_ICCV2025_tang2025hip}
& 90.56 & \underline{85.90} & 0.50 & 1.07 & 1.74 & 1.10 & \textbf{0.05} & \textbf{0.16} & \textbf{0.27} & \textbf{0.16} & \underline{0.6618} & \textbf{0.1669} & \underline{0.7534} \\

\midrule

\rowcolor[HTML]{EEF5FF}
HiP-AD + DPT + SM
& \textbf{95.95} & \textbf{86.57} & \underline{0.46} & \textbf{0.95} & \textbf{1.55} & \underline{0.99} & \textbf{0.05} & \textbf{0.16} & \underline{0.28} & \textbf{0.16} & \textbf{0.6736} & \underline{0.1749} & \textbf{0.7554} \\
\bottomrule
\end{tabular}%
}
\vspace{-10pt}
\end{table}

Therefore, we conduct experiments to examine whether DPT can effectively alleviate the conflict between perception and planning tasks even in E2E models with such parallel architectures. As the backbone model, we adopt HiP-AD~\cite{supp-E2E_4_ICCV2025_tang2025hip}, a recent model with a parallel architecture. For DPT on HiP-AD, we reorganize the original training objectives into two groups. The perception-related group includes losses for 3D object detection, map prediction, motion prediction, and auxiliary dense depth estimation, whereas the planning-related group includes losses for ego-status estimation and trajectory planning. We apply DPT based on this grouping to mitigate the conflict between perception and planning in the shared representation, and then integrate the resulting task-specific parameters into a single HiP-AD model via sparse merging for evaluation.

Following the HiP-AD setup, we conduct experiments on Bench2Drive~\cite{supp-E2E_36_NeurIPS2024_jia2024bench2drive} and evaluate on the Dev10 test benchmark introduced in DriveTransformer~\cite{supp-E2E_25_ICLR2025_jia2025drivetransformer}. For planning evaluation, we use Route Completion (RC) and Driving Score (DS) as closed-loop metrics. For open-loop evaluation, we use L2 distance and collision rate for planning, and mAP, mATE, and NDS for perception. As baselines, we evaluate UniAD~\cite{supp-E2E_42_old_hu2023planning_endtoend7}, VAD~\cite{supp-E2E_39_old_jiang2023vad_endtoend3}, DriveTransformer~\cite{supp-E2E_25_ICLR2025_jia2025drivetransformer}, and HiP-AD using the official checkpoints and configurations provided on GitHub, and compare them with HiP-AD enhanced with DPT and Sparse Merging. The quantitative comparison is summarized in Tab.~\ref{tab:E2E_metric}. The results suggest that our method is applicable to end-to-end autonomous driving and can improve both closed-loop and open-loop performance on HiP-AD.

\section{Future Direction}
\label{sub:future_direction}

As a future direction, DPT can be extended to more general multi-task settings with a larger number of tasks.
While this work mainly focuses on prediction and planning, the supplementary analyses further examine sub-task decomposition and end-to-end autonomous driving. Broader unified models with more diverse tasks may require more principled designs of task grouping and parameter allocation. In addition, although we provide sensitivity analyses for several DPT settings, adaptive selection of DPT hyperparameters, such as the allocation ratio, mask ratio, and allocation order, remains an important direction. We expect these directions to further improve the applicability of DPT across broader unified model settings.

\end{document}